%% file: iclr2024_conference.tex
\documentclass{article} % For LaTeX2e
\usepackage{iclr2024_conference,times}
\iclrfinalcopy
\input{preamble}

\usepackage[utf8]{inputenc} % allow utf-8 input
\usepackage[T1]{fontenc}    % use 8-bit T1 fonts
\usepackage[colorlinks=true, 
    linkcolor=violet, 
    citecolor=teal, 
    filecolor=magenta, 
    urlcolor=teal,]{hyperref}
\usepackage{url}            % simple URL typesetting
\usepackage{booktabs}       % professional-quality tables
\usepackage{amsfonts}       % blackboard math symbols
\usepackage{nicefrac}       % compact symbols for 1/2, etc.
\usepackage{microtype}      % microtypography
\usepackage{xcolor}  
\usepackage{graphicx}% colors
\usepackage{subcaption}
\usepackage{array}
\usepackage{tabularx}
\usepackage{wrapfig}
\usepackage{amssymb}

\title{AdjointDPM: Adjoint Sensitivity Method for Gradient Backpropagation of Diffusion Probabilistic Models}

\author{%
  Jiachun Pan\thanks{Equal contribution. This work was completed during Jiachun Pan's internship at ByteDance.} \\ 
    National University of Singapore\\ \texttt{pan.jc@nus.edu.sg} \\
  \And
  Jun Hao Liew \\
  ByteDance \\
  \texttt{junhao.liew@bytedance.com} \\
  \And
  Vincent Y. F. Tan \\
  National University of Singapore \\
  \texttt{vtan@nus.edu.sg} \\
  \And
  Jiashi Feng \\
  ByteDance \\
  \texttt{jshfeng@bytedance.com} \\
  \And
  Hanshu Yan\footnotemark[1] ~\thanks{Project Lead.} \\
  ByteDance \\
  \texttt{hanshu.yan@bytedance.com} \\
}

\newcommand{\eg}{\textit{e.g.}}
\newcommand{\ie}{\textit{i.e.}}

\newcommand{\invtoken}{\textbf{\#}}
\newcommand{\cinvtoken}{\textcolor{magenta}{\textbf{\#}}}

\begin{document}

\maketitle

\begin{abstract}
% Existing customization methods require access to multiple reference examples to align pre-trained diffusion probabilistic models (DPMs) with user-provided concepts. 
This paper considers a ubiquitous problem underlying several applications of DPMs, \ie, 
% the customization and guidance applications is how to optimize the diffusion models' parameters so that the generated contents satisfy certain properties. 
% This paper aims to address the challenge of 
optimizing the parameters of DPMs when the objective is a differentiable metric defined on the generated contents. 
Since the sampling procedure of DPMs involves recursive calls to the denoising UNet, na\"ive gradient backpropagation requires storing the intermediate states of all iterations, resulting in extremely high memory consumption. 
To overcome this issue, we propose a novel method AdjointDPM, which first generates new samples from diffusion models by solving the corresponding probability-flow ODEs. It then uses the adjoint sensitivity method to backpropagate the gradients of the loss to the models' parameters (including conditioning signals, network weights, and initial noises) by solving another augmented ODE. 
To reduce numerical errors in both the forward generation and gradient backpropagation processes, we further reparameterize the probability-flow ODE and augmented ODE as simple non-stiff ODEs using exponential integration. 
AdjointDPM can effectively compute the gradients of all types of parameters in DPMs, including the network weights, conditioning text prompts, and noisy states.
Finally, we demonstrate the effectiveness of AdjointDPM on several interesting tasks: guided generation via modifying sampling trajectories, finetuning DPM weights for stylization, and converting visual effects into text embeddings.\footnote{
Github link for codes: \url{https://github.com/HanshuYAN/AdjointDPM.git}
}
\end{abstract}

\section{Introduction}
Diffusion Probabilistic Models (DPMs) constitute a family of generative models that diffuse data distributions into white Gaussian noise and then revert the stochastic diffusion process to synthesize new contents \citep{ho_denoising_2020, song2020score}. DPM-based methods have recently achieved state-of-the-art performances in generating various types of contents, such as images \citep{saharia2022photorealistic, rombach_high-resolution_2022, ramesh_hierarchical_2022}, videos \citep{blattmann_align_2023, zhou_magicvideo_2022, ho_imagen_2022}, and audio data~\citep{liu_audioldm_2023, schneider_archisound_2023}. To promote the development of downstream tasks, several pre-trained high-performance models, such as Stable Diffusion (SD) \citep{rombach_high-resolution_2022}, have been made publicly available. Based on these public large-scale models, researchers have developed many algorithms for creative applications \citep{gal_image_2022, daras_multiresolution_2022, kawar_imagic_2023, ruiz_dreambooth_2022, fei_gradient-free_2023,wen_hard_2023, molad_dreamix_2023}. For example, a line of customization algorithms of DPMs, such as Textural-Inversion~\citep{gal_image_2022} and DreamBooth~\citep{ruiz_dreambooth_2022}, have been proposed to adapt DPMs for generating images/videos that share certain styles or identities. Researchers also have proposed some guidance algorithms~\citep{bansal_universal_2023, ho_classifier-free_2022} to make the generation process more controllable.

A ubiquitous problem in customization and guidance applications is the  optimization of the diffusion models' parameters so that the final generated contents satisfy certain properties. 
For example, to customize models for certain styles, we need to optimize the model weights to minimize the style distance between the generated images and the reference. Alternatively, concerning the guidance of sampling, we need to adjust the intermediate noisy states via the gradients of the guidance loss computed on the final generated data. Generally, the parameters to optimize include the conditional text embeddings, the network weights, and the noisy states, as they all affect the sampling trajectories. We formulate this optimization problem as follows. Denote the DPM as $\Phi(\cdot, \cdot, \epsilon_{\theta})$, which generates samples by iteratively calling a function $\epsilon_{\theta}$. The desired properties can be defined by the loss function $L(\cdot)$ computed based on the generated contents $\bx_0=\Phi(\bx_T, c, \epsilon_{\theta})$. We aim to minimize the loss by optimizing the   variables $\psi$, including the weights $\theta$, conditioning signals $c$, or initial noise $\bx_T$, i.e.,
\begin{align}
\label{eqn:opt}
     \min_{\psi \in \{\bx_T, c, \theta \}} L(\Phi(\bx_T, c, \bepsilon_{\theta})).
\end{align}

To solve the optimization problem~\eqref{eqn:opt}, an effective backpropagation (BP) technique is required to compute the gradient of the loss function $L(\bx_0)$ with respect to the optimization variables. \citet{song2020score} showed that the DPM sampling process is equivalent to solving a probability-flow ODE. Thus, many efficient sampling methods have been developed using adaptive ODE solvers. The sampling process involves recursive calls to the denoising UNet $\epsilon_{\theta}(\bx_t, t, c)$ for multiple iterations. 
Using na\"ive gradient BP requires intermediate state storage for all iterations, resulting in significant GPU memory consumption. To overcome this problem, we propose AdjointDPM, a novel gradient BP technique based on the adjoint sensitivity method~\citep{chen2018neural}. AdjointDPM computes the gradient by solving a backward ODE that only needs to store the intermediate state at the time point of function evaluation, resulting in constant memory usage.
Moreover, we reparameterize the diffusion generation process to a simple non-stiff ODE using exponential integration, which helps reduce discretization errors in both the forward and reverse processes of gradient computation.

% Solving the optimization problem~\eqref{eqn:opt}, we can perform several interesting applications. For instance, one may optimize a plug-in conditioning text token or slightly fine-tune the model's weights for image style transfer. The style is defined by the Gram matrix \citep{gatys_neural_2015} computed from a single reference image's features. Alternatively, one may evaluate the security of a generation system by checking whether there exists an adversarial initial noise $\bx_T$ that can mislead the content moderation mechanism. The objective is the classification score of $\Phi(\bx_T, c, \epsilon_{\theta})$ by the NSFW (\textbf{n}ot \textbf{s}afe \textbf{f}or \textbf{w}ork) filter. 

We evaluate the effectiveness of AdjointDPM by applying it to several interesting tasks,  
involving optimizing the initial/intermediate noisy states, network weights, and conditioning text prompts, respectively. 
% In all the tasks, the metrics are all manually designed, \eg, a classifier, the distance between a stylized Gram Matrix. 
% The results demonstrate the flexibility and general applicability of AdjointDPM:
\textcolor{teal}{1)} Guided sampling. Under the supervision of fine-grained vision classifiers, AdjointDPM can guide the Stable Diffusion to synthesize images of certain breeds of animals. 
\textcolor{teal}{2)} Security auditing of image generation systems. AdjointDPM successfully finds a set of initial noise whose corresponding output images contain NSFW (\textbf{n}ot \textbf{s}afe \textbf{f}or \textbf{w}ork) content, but can sneakily bypass the moderation filters. This triggers \textit{an alert about the potential security issues of existing AI generation systems}.
% \textcolor{teal}{2)} AdjointDPM manages to optimize a unique special text embedding \invtoken ~that can control a Stable Diffusion model to synthesize images with certain visual effects (\eg, bokeh and relighting).
\textcolor{teal}{3)} Stylization via a single reference image. AdjointDPM can finetune a Stable Diffusion model for stylization defined by the Gram Matrix \citep{gatys_neural_2015} of a reference image. The stylizing capability of the fine-tuned model can generalize to different objects.
All the empirical results demonstrate the flexibility and general applicability of AdjointDPM. We summarize the main contributions of this paper as follows:
\begin{enumerate}
    \item We propose a novel gradient backpropagation method of DPMs by applying the adjoint sensitivity method to the sampling process of diffusion models. 
    \item To the  best of our knowledge, AdjointDPM is the first general gradient backpropagation method that can be used for all types of parameters of DPMs, including network weights, conditioning text prompts, and intermediate noisy states.
    \item AdjointDPM can be used for several creative applications and outperforms other baselines.
\end{enumerate}

\section{Background}
\label{sec:background}

\subsection{Probability-flow ODEs corresponding to DPMs}
\label{subsec:diff}
% In this work, we mainly consider the ODE form of diffusion models.

The framework of diffusion probabilistic models involves gradually diffusing the complex target data distribution to a simple noise distribution, such as white Gaussian, and solving the corresponding reverse process to generate new samples~\citep{ho_denoising_2020, song2020score}. Both the diffusion and denoising processes can be characterized by temporally continuous stochastic differential equations (SDEs)~\citep{song2020score}. \citet{song2020score} derive deterministic processes (probability-flow ODEs) that are equivalent to the stochastic diffusion and denoising processes in the sense of marginal probability densities for all the time steps. 

Let $q_0$ denote the unknown $d$-dimensional data distribution. \citet{song2020score} formulated the forward diffusion process $\{\bx(t)\}_{t\in[0,T]}$ as  follows
\begin{align}
\label{eqn:forw}
    \rmd \bx_t=f (t)\bx_t\, \rmd t+g(t)\, \rmd \bw_t,\quad \bx_0\sim q_0(\bx),\quad t\in[0,T],
\end{align}
where $\bx_t$ denotes the state at time $t$ and $\bw_t$ is the standard Wiener process. $f (t)\bx_t$ is a vector-valued function called \emph{drift} coefficient and $g(t)$ is a scalar function known as \emph{diffusion} coefficient. For the forward diffusion process, it is common to adopt the conditional probability as $q_{0t}(\bx_t|\bx_0)=\calN(\bx_t|\alpha_t\bx_0,\sigma_t^2\bI)$,
and the marginal distribution of $\bx_T$ to be approximately a standard Gaussian. For sampling, we have a corresponding reverse process as 
\begin{align}
\label{eqn:reverse}
    \rmd\bx_t=[f (t)\bx_t-g(t)^2\nabla_{\bx_t}\log q_{t}(\bx_t)]\, \rmd t+g(t)\, \rmd \bw_t,\quad \bx_T\sim \calN(0,\sigma_T^2\bI), \quad t\in[0,T].
\end{align}

In Eqn.~\eqref{eqn:reverse}, the term $\nabla_{\bx}\log q_t(\bx)$, is known as the \emph{score function}. We can train a neural network $\bepsilon_{\theta}(\bx_t,t)$ to estimate $-\sigma_t\nabla_{\bx}\log q_t(\bx_t)$ via denoising score matching.
% matching $\bepsilon_{\theta}(\bx_t,t)$ and $-\sigma_t\nabla_{\bx}\log q_t(\bx_t)$ with the objective function as follows:
% \begin{align}
% \label{eqn:loss}
%     \calL_{\mathrm{SM}}(\theta):=\bbE_t\big\{\lambda(t)\bbE_{\bx_0}\bbE_{\bepsilon}[\|\bepsilon_{\theta}(\bx_t,t)-\bepsilon\|_2^2]\big\},
% \end{align}
% where $\lambda(\cdot)>0$ is a weighting function, $\bepsilon\sim\calN(\bepsilon|0,\bI)$, and $\bx_t=\alpha_t\bx_0+\sigma_t\bepsilon$.
As discussed in \cite{song2020score}, there exists a corresponding deterministic process whose trajectory shares the same set of marginal probability densities $\{q_t(\bx)\}_{t=0}^T$ as the SDE~\eqref{eqn:reverse}. 
% For a well-trained model $\bepsilon_{\theta}(\bx_t,t)$, 
The form of this deterministic probability-flow ODE is shown in ~\eqref{eqn:ode2}. We can generate new samples by solving Eqn.~\eqref{eqn:ode2} from $T$ to $0$ with initial sample $\bx_T$ drawn from $ \calN(0,\sigma_T^2\bI)$. 
% , and the  form of the ODE  is 
% \begin{align}
% \label{eqn:ode}
%     \rmd\bx =\left[f (t)\bx_t-\frac{1}{2}g(t)^2\nabla_{\bx}\log q_{t}(\bx)\right]\rmd t.
% \end{align}
% Thus, we can use the well-trained neural network $\bepsilon_{\theta}(\bx_t,t)$ to generate new samples by solving Eqn.~\eqref{eqn:ode2} from $T$ to $0$ with initial sample $\bx_T$ drawn from $ \calN(0,\sigma_T^2\bI)$. 
% This is the ODE form of unconditional generation. 
\begin{align}
\label{eqn:ode2}
    \rmd\bx =\left[f (t)\bx_t+\frac{g(t)^2}{2\sigma_t}\bepsilon_{\theta}(\bx_t,t)\right]\rmd t
\end{align}
For conditional sampling, classifier-free guidance (CFG)~\citep{ho_classifier-free_2022} has been widely used in various tasks for improving the sample quality, including text-to-image, image-to-image, class-to-image generation~\citep{saharia2022photorealistic,dhariwal2021diffusion,nichol2021glide}. 
% During training, one may simultaneously train unconditional and conditional prediction models with the same parameterized model $\bepsilon_{\theta}(\bx_t,t,c)$. The unconditional mode is triggered by setting $c$ to be a fixed special placeholder $\varnothing$. Hence, the sampling model becomes
 % \begin{align*}
 %     \tilde{\bepsilon}_{\theta}(\bx_t,t,c):= s\cdot \bepsilon_{\theta}(\bx_t,t,c)+(1-s)\cdot \bepsilon_{\theta}(\bx_t,t,\varnothing),
 % \end{align*}
We can use CFG to generate new samples by solving Eqn.~\eqref{eqn:ode2} and replacing $\bepsilon_{\theta}(\bx_t,t)$ with $\tilde{\bepsilon}_{\theta}(\bx_t,t,c)$.
 \begin{align*}
     \tilde{\bepsilon}_{\theta}(\bx_t,t,c):= s\cdot \bepsilon_{\theta}(\bx_t,t,c)+(1-s)\cdot \bepsilon_{\theta}(\bx_t,t,\varnothing),
 \end{align*}

% This paper mainly focuses on the optimization problem based on the generated samples of DPMs solved by probability-flow ODEs. Denote the generation mapping of a probability-flow ODE as $\Phi$ and  our generated samples as 
% $$\bx_0=\Phi(\bx_T,c,\bepsilon_{\theta})=\bx_T+\int_{T}^0 \left[f (t)\bx_t+\frac{g(t)^2}{2\sigma_t}\tilde{\bepsilon}_{\theta}(\bx_t,t, c)\right] ~\rmd t.$$
% The output samples are determined by the initial noise $\bx_T$, the conditioning signal $c$, and the denoising models $\bepsilon_{\theta}$. Thus, if we want to customize the generated contents based on a loss function on $\bx_0$, we have three options---tuning the initial noise $\bx_T$, the conditional variable $c$, or the weights $\theta$ by backpropagating the loss. 

\subsection{Adjoint Sensitivity Methods for Neural ODEs}
\label{subsec:adjoint}
Considering a neural ODE model $$\frac{\rmd\bx}{\rmd t} =\bs(\bx_t, t, \theta),$$
the output $\bx_0 = \bx_T+\int_{T}^0 \bs(\bx_t, t, \theta) ~\rmd t$.
We aim to optimize the input $\bx_T$ or the weights $\theta$ by minimizing a loss $L$ defined on the output $\bx_0$. 
% Thus, we need to compute the gradients $\{\frac{\partial L}{\partial \bx_T}, \frac{\partial L}{\partial \theta} \}$. 
Regarding $\frac{\partial L}{\partial \bx_T}$, \citet{chen2018neural} introduced adjoint state $\ba(t)=\frac{\partial L}{\partial \bx_t}$, which represents how the loss w.r.t the state $\bx_t$ at any time $t$. 
The dynamics of $\ba(t)$ are given by another ODE, 
\begin{align}
    \label{eqn:adjoint}
    \frac{\rmd \ba(t)}{\rmd t}=-\ba(t)^T \frac{\partial \bs(\bx_t, t, \theta)}{\partial \bx_t},
\end{align} 
which can be thought of as the instantaneous analog of the chain rule. Since $\frac{\partial L}{\partial \bx_0}$ is known, we can compute $\frac{\partial L}{\partial \bx_T}$ by solving the initial value problem (IVP) backwards in  time $T$ to $0$ of ODE in~\eqref{eqn:adjoint}. Similarly, for $\theta$, we can regard them as a part of the augmented state:
\begin{align*}
    \frac{\rmd}{\rmd t} \left[
    \begin{matrix}
    \bx,
    \theta,
    t
    \end{matrix}\right](t): =
    \left[\begin{matrix}
        \bs(\bx_t, t, \theta),
        \mathbf{0},
        1
    \end{matrix}\right].
\end{align*}
% Noted that we omit the parameter $c$ here because both $c$ and $\theta$ serve as the \textit{control} of the dynamics and have the same functionality in the ODE. One can use the following method to compute $\frac{\partial L}{\partial c}$ by simply replacing $\theta$ to $c$.
The corresponding adjoint state to this augmented state are $\ba_{\text{aug}}(t):= \left[\begin{matrix}
        \ba(t),
        \ba_{\theta}(t),
        \ba_{t}(t)
    \end{matrix}\right]$, where $\ba_{\theta}:=\frac{\partial L}{\partial \theta}$ and $\ba_t:=\frac{\partial L}{\partial t}$.  
% \begin{align*}
%     \ba_{\text{aug}}(t):= \left[\begin{matrix}
%         \ba(t),
%         \ba_{\theta}(t),
%         \ba_{t}(t)
%     \end{matrix}\right], \quad
%     \ba_{\theta}:=\frac{\partial L}{\partial \theta}, \quad \mbox{and}\quad \ba_t:=\frac{\partial L}{\partial t}.
% \end{align*}
The augmented adjoint state $\ba_{\text{aug}}$ is governed by:
\begin{align}
\label{eqn:augode}
    \frac{\rmd \ba_{\text{aug}}}{\rmd t}=- \begin{bmatrix}\ba\frac{\partial\bs}{\partial \bx},\ba\frac{\partial\bs}{\partial \theta},\ba\frac{\partial\bs}{\partial t}\end{bmatrix}.
\end{align}

By solving the IVP from time $T$ to $0$ of Eqn.~\eqref{eqn:augode}, we  obtain the gradients of $L$ w.r.t.\ $ \{\bx_t, \theta, t\}$. 
% Thus, the gradients can be calculated altogether during a single call of the ODE solver with augmented dynamics.
The explicit algorithm~\citep{chen2018neural} is shown in Algorithm~\ref{alg:node}. 
\begin{algorithm}
\caption{Reverse-mode derivative of an ODE initial value problem}
\label{alg:node}
\textbf{Input:} Dynamics parameter $\theta$, start time $t_0$, end time $t_1$, final state $\bx_{t_1}$, loss gradient $\partial L/\partial \bx_{t_1}$.
\begin{algorithmic}
\Statex $a(t_1)=\frac{\partial L}{\partial \bx_{t_1}}$, $a_{\theta}(t_1)=\textbf{0}$, $z_0=[\bx_{t_1}, a(t_1), a_{\theta}(t_1)]$ \Comment{Define initial augmented state.}
\Statex \textbf{def} AugDynamics($[\bx_t, \ba_t, \cdot], t,\theta$) \Comment{Define dynamics on augmented state.}\\
        % $\qquad\textbf{return}$ $[\bs(\bx_t, t, \theta, c), -\ba_t^T \frac{\partial \bs}{\partial \bx}, -\ba_t^T \frac{\partial \bs}{\partial \theta}
        % % -\ba_t^T \frac{\partial \bs}{\partial t}
        % ]$ \Comment{Concatenate time-derivatives}
        $\qquad\textbf{return}$ $[\bs(\bx_t, t, \theta), -\ba_t^T \frac{\partial \bs}{\partial \bx}, -\ba_t^T \frac{\partial \bs}{\partial \theta}
        % -\ba_t^T \frac{\partial \bs}{\partial t}
        ]$ \Comment{Concatenate time-derivatives}

\Statex $[\bx_{t_0},  \frac{\partial L}{\partial \bx_{t_0}}, \frac{\partial L}{\partial \theta}
% \frac{\partial L}{\partial t_0}
]=\mathrm{ODESolve}(z_0,\mathrm{AugDynamics}, t_1, t_0, \theta )$ \Comment{Solve reverse-time ODE}
\end{algorithmic}
% \textbf{Return:} $[\frac{\partial L}{\partial \bx_{t_0}}, \frac{\partial L}{\partial \theta},\frac{\partial L}{\partial t_0},\frac{\partial L}{\partial t_1}]$ \Comment{Return all gradients}
\textbf{Return:} $[\frac{\partial L}{\partial \bx_{t_0}}, \frac{\partial L}{\partial \theta}]$ \Comment{Return gradients}
\end{algorithm}

\section{Adjoint Sensitivity Methods for Diffusion Probabilistic Models}
% \section{Improvements of Adjoint Sensitivity Methods on Diffusion Models}
\label{sec:improve}

In this section, we develop the AdjointDPM for gradient backpropagation in diffusion models based on the adjoint sensitivity methods from the neural ODE domain. 
When optimizing the model's parameters $\bx_T$ or $\theta$ (including the conditioning $c$), AdjointDPM first generates new samples via the forward probability-flow ODE~\eqref{eqn:ode2}. Through applying the adjoint sensitivity method, we then write out and solve the backward adjoint ODE~\eqref{eqn:augode} to compute the gradients of loss with respect to the parameters. One can apply any general-purpose numerical ODE solver, such as Euler--Maruyama and Runge--Kutta methods~\citep{atkinson2011numerical}, for solving the ODE. To further improve the efficiency of the vanilla adjoint sensitivity methods, we exploit the semi-linear structure of the diffusion ODE functions~\eqref{eqn:ode2}, which has been used in several existing works for accelerating DPM samplers \citep{lu2022dpm,lu2022dpm2,karraselucidating,zhang2022fast}, and reparameterize the forward and backward ODEs as simple non-stiff ones.

% To improve the sampling efficiency, researchers have proposed several advanced samplers \citep{lu2022dpm,lu2022dpm2,karraselucidating,zhang2022fast} by considering the semi-linear structure of diffusion ODE functions~\eqref{eqn:ode2}. In this section, we improve the adjoint sensitivity methods by following similar ideas. 

\subsection{Applying Adjoint Methods to Probability-flow ODEs}
Sampling from DPMs, we obtain the generated data $\bx_0 = \bx_T+\int_{T}^0 \bs(\bx_t, t, \theta, c) ~\rmd t$, where 
\begin{align}
    \label{eqn:prob-ode-dyn}
    \bs(\bx_t, t, \theta, c) = f (t)\bx_t+\frac{g(t)^2}{2\sigma_t}\tilde{\bepsilon}_{\theta}(\bx_t,t, c). 
\end{align}
% $\bs(\bx_t, t, \theta, c) = f (t)\bx_t+\frac{g(t)^2}{2\sigma_t}\tilde{\bepsilon}_{\theta}(\bx_t,t, c)$. 
Concerning the customization or guidance tasks, we aim to minimize a loss $L$ defined on $\bx_0$, such as the stylization loss or semantic scores. We plug the equation \eqref{eqn:prob-ode-dyn} into the augmented adjoint ODE \eqref{eqn:adjoint}, and obtain the reverse ODE function in Algorithm~\ref{alg:node} as:
\begin{align}
\label{eqn:reverseode}
    \rmd \left[\begin{matrix}
        \bx_{t}\vspace{.3em}\\
        \frac{\partial L}{\partial \bx_t}\vspace{.3em}\\
        \frac{\partial L}{\partial \theta}\vspace{.3em}\\
        \frac{\partial L}{\partial t}
    \end{matrix}\right]=-\left[\begin{matrix}
    -f (t)\bx_t-\frac{g(t)^2}{2\sigma_t}\tilde{\bepsilon}_{\theta}(\bx_t,t,c)\\
    f(t) \frac{\partial L}{\partial \bx_t} + \frac{\partial L}{\partial \bx_t} \frac{g(t)^2}{2\sigma_t}\frac{\partial \tilde{\bepsilon}_{\theta}(\bx_t,t,c)}{\partial \bx_t}\\
    \frac{\partial L}{\partial \bx_t} \frac{g(t)^2}{2\sigma_t}\frac{\partial \tilde{\bepsilon}_{\theta}(\bx_t,t,c)}{\partial \theta}\\
    \frac{\rmd f(t)}{\rmd t} \frac{\partial L}{\partial \bx_t} \bx_t + \frac{\partial L}{\partial \bx_t} \frac{\partial [{g(t)^2}/{2\sigma_t}\tilde{\bepsilon}_{\theta}(\bx_t,t,c)]}{\partial t}
    \end{matrix}\right ]\rmd t.
\end{align}
We observe that the ODEs governing $\bx_t$ and $\frac{\partial L}{\partial \bx_t}$ both contain linear and nonlinear parts. If we directly use off-the-shelf numerical solvers on Eqn.~\eqref{eqn:reverseode}, it causes discretization errors of both the linear and nonlinear terms. To avoid this, in Section \ref{subsec:reparameterization}, we exploit the semi-linear structure of the probability-flow ODE to better control the discretization error for each step. Thus, we are allowed to use a smaller number of steps for generating samples of comparable quality. 
% The efficiency of the vanilla adjoint sensitivity methods will also be improved.
% Next, we will discuss how to improve the efficiency of the vanilla adjoint sensitivity methods.

\subsection{Exponential Integration and Reparameterization}
\label{subsec:reparameterization}
We use the \emph{exponential integration} to transform the ODE~\eqref{eqn:ode2} into a simple non-stiff ODE. We multiply an integrating factor $\exp(-\int_{0}^{t} f(\tau)\rmd \tau)$ on both sides of Eqn.~\eqref{eqn:ode2} and obtain 
\begin{align*}
    \frac{\rmd e^{-\int_{0}^{t} f(\tau)\rmd \tau} \bx_t }{\rmd t}=e^{-\int_{0}^{t} f(\tau)\rmd \tau}\frac{g(t)^2}{2\sigma_t}\tilde{\bepsilon}_{\theta}(\bx_t,t,c).
\end{align*}
Let $\by_t$ denote $e^{-\int_{0}^{t} f(\tau)\rmd \tau} \bx_t$, then we have 
\begin{align}
\label{eqn:ode_yt}
    \frac{\rmd \by_t }{\rmd t}=e^{-\int_{0}^{t} f(\tau)\rmd \tau}\frac{g(t)^2}{2\sigma_t}\tilde{\bepsilon}_{\theta}\left(e^{\int_{0}^{t} f(\tau)\rmd \tau}\by_t,t,c\right).
\end{align}
We introduce a variable $\rho = \gamma(t)$ and $\frac{\rmd \gamma}{\rmd t}= e^{-\int_{0}^{t} f(\tau)\rmd \tau}\frac{g(t)^2}{2\sigma_t}$.
In diffusion models, $\gamma(t)$ usually monotonically increases when $t$ increases from $0$ to $T$.
% the choice of $e^{-\int_{0}^{t} f(\tau)\rmd \tau}\frac{g(t)^2}{2\sigma_t}$ is usually a monotone function with respect to $t$. 
For example, when we choose $f(t)=\frac{\rmd \log \alpha}{\rmd t}$ and $g^2(t)=\frac{\rmd \sigma_t^2}{\rmd t}-2\frac{\rmd \log \alpha}{\rmd t}\sigma_t^2$ in VP-SDE~\citep{song2020score}, we have $\gamma(t) = \alpha_0 \frac{\sigma_t}{\alpha_t}-\sigma_0$.
% , which monotonically increases when $t$ increases from $0$ to $T$. 
Thus, a bijective mapping exists between $\rho $ and $t$, and we can reparameterize \eqref{eqn:ode_yt} as:
\begin{align}
\label{eqn:newode}
    \frac{\rmd \by }{\rmd \rho}=\tilde{\bepsilon}_{\theta}\Big(e^{\int_{0}^{\gamma^{-1}(\rho)} f(\tau)\rmd \tau}\by,\gamma^{-1}(\rho),c\Big).
\end{align}
% After reparameterizing the diffusion ODE \eqref{eqn:ode2} as \eqref{eqn:newode}, 
We also reparameterize the reverse ODE function in Algorithm~\ref{alg:node} as follows
\begin{align}
\label{eqn:reparareverse}
    \rmd \left[\begin{matrix}
        \by\vspace{.3em}\\
        \frac{\partial L}{\partial \by} \vspace{.3em}\\
        \frac{\partial L}{\partial \theta}\vspace{.3em} \\
        \frac{\partial L}{\partial \rho}
    \end{matrix}\right]=-\left[\begin{matrix}
    -\tilde{\bepsilon}_{\theta}\big(e^{\int_{0}^{\gamma^{-1}(\rho)} f(\tau)\rmd \tau}\by,\gamma^{-1}(\rho),c\big)\\
     \frac{\partial L}{\partial \by} \frac{\partial \tilde{\bepsilon}_{\theta}\big(e^{\int_{0}^{\gamma^{-1}(\rho)} f(\tau)\rmd \tau}\by,\gamma^{-1}(\rho),c\big)}{\partial \by}\\
    \frac{\partial L}{\partial \by} \frac{\partial \tilde{\bepsilon}_{\theta}\big(e^{\int_{0}^{\gamma^{-1}(\rho)} f(\tau)\rmd \tau}\by,\gamma^{-1}(\rho),c\big)}{\partial \theta}\\
     \frac{\partial L}{\partial \by} \frac{\partial \tilde{\bepsilon}_{\theta}\big(e^{\int_{0}^{\gamma^{-1}(\rho)} f(\tau)\rmd \tau}\by,\gamma^{-1}(\rho),c\big)}{\partial \rho}
    \end{matrix}\right]\rmd \rho.
\end{align}
Now instead of solving Eqn.~\eqref{eqn:ode2} and Eqn.~\eqref{eqn:reverseode},
we use off-the-shelf numerical ODE solvers to solve Eqn.~\eqref{eqn:newode} and Eqn.~\eqref{eqn:reparareverse}. This method is termed AdjointDPM. Implementation details are provided in Appendix \ref{apdx:implementation}.

\subsection{Error Control} 
Here, we first show that the exact solutions of the reparameterzied ODEs are equivalent to the original ones. For the equation in the first row of Eqn.~\eqref{eqn:reparareverse}, its exact solution is:
\begin{align}
\label{eqn:exactsol2}
    \by_{\rho(t)} = \by_{\rho(s)} +\int_{\rho(s)}^{\rho(t)} \tilde{\bepsilon}_{\theta}\Big(e^{\int_{0}^{\gamma^{-1}(\rho)} f(\tau)\rmd \tau}\by,\gamma^{-1}(\rho),c\Big)\, \rmd \rho.
\end{align}
We can rewrite it as $e^{-\int_{0}^{t} f(\tau)\rmd \tau} \bx_t = e^{-\int_{0}^{s} f(\tau)\rmd \tau} \bx_s +\int_{s}^{t} \frac{\rmd \rho}{\rmd \tau } \tilde{\bepsilon}_{\theta}(\bx_{\tau},\tau,c) \, \rmd \tau$. Then, we have
\begin{align*}
    % &e^{-\int_{0}^{t} f(\tau)\rmd \tau} \bx_t = e^{-\int_{0}^{s} f(\tau)\rmd \tau} \bx_s +\int_{s}^{t} \frac{\rmd \rho}{\rmd \tau } \tilde{\bepsilon}_{\theta}(\bx_{\tau},\tau,c) \, \rmd \tau \\
    % \Rightarrow 
    \quad & \bx_t = e^{\int_s^t f(\tau)\rmd \tau} \bx_s + \int_{s}^t e^{\int_{\tau}^t f(r)\rmd r} \frac{g(\tau)^2}{2\sigma_{\tau}} \tilde{\bepsilon}_{\theta}(\bx_{\tau},\tau,c) \, \rmd \tau,
\end{align*}
which is equivalent to the exact solution of the equation in the first row of Eqn.~\eqref{eqn:reverseode}.
Similarly, for other equations in~\eqref{eqn:reparareverse}, their exact solutions are also equivalent to the solutions in ~\eqref{eqn:reverseode}.
% for the second equation in~\eqref{eqn:reparareverse}, its exact solution is
% \begin{align}
% \label{eqn:exactsol3}
%      &\frac{\partial L }{\partial \by_{\rho(t)}} = \frac{\partial L }{\partial \by_{\rho(s)}} - \int_{\rho(s)}^{\rho(t)}\frac{\partial L }{\partial \by_{\rho}}\frac{\partial\tilde{\bepsilon}_{\theta}\big(e^{\int_{0}^{\gamma^{-1}(\rho)} f(\tau)\rmd \tau}\by,\gamma^{-1}(\rho),c\big)}{\partial \by_{\rho}}\, \rmd \rho\\
%      \Rightarrow \quad &\frac{\partial L }{\partial \bx_t} = e^{\int_{t}^{s} f(\tau)\rmd \tau}\frac{\partial L }{\partial \bx_s}- \int_{s}^t \frac{\partial L }{\partial \bx_{\tau}}e^{\int_{t}^{\tau}f(r)\rmd r}\frac{g(\tau)^2}{2\sigma_{\tau}}\frac{\partial \tilde{\bepsilon}_{\theta}(\bx_{\tau},\tau,c)}{\partial \bx_{\tau}}\, \rmd \tau\notag,
% \end{align}
% which is also equivalent to the solution of the second equation in \eqref{eqn:reverseode}. 
% \textcolor{blue}{////////////////explain why the numerical error can be better controlled in this parameterization type}\\
% Then..... \\
% Then, the analyses on the discretization errors when we solve Eqn~\eqref{eqn:newode} and Eqn~\eqref{eqn:reparareverse} using numerical ODE solvers are shown in Proposition~\ref{pro:error}. The proof of Proposition~\ref{pro:error} will be shown in the Appendix. The main idea of proof is that 
Thus, when we numerically solve non-stiff ODEs in Eqns~\eqref{eqn:newode} and~\eqref{eqn:reparareverse}, there are only discretization errors for nonlinear functions and the closed form of integration of linear parts have been solved exactly without any numerical approximation.

% {\color{blue}
% \begin{proposition}
% \label{pro:error}
%     When we solve Eqn~\eqref{eqn:newode} and Eqn~\eqref{eqn:reparareverse} using the off-the-shelf numerical ODE solver, the discretization error of the linear part will not be introduced.
% \end{proposition}
% }
In summary, we reformulate the forward and reverse ODE functions 
% in this section 
and show that by using off-the-shelf numerical ODE solvers on the reparameterized ODEs, 
AdjointDPM does not introduce discretization eerror to the linear part.
% the discretization error of the linear part will not be introduced by AdjointDPM. 
In Section~\ref{subsec:fid}, we experimentally compare the FID of generated images by solving  Eqn.~\eqref{eqn:ode2} and Eqn.~\eqref{eqn:newode} with the same number of network function evaluations (NFE). The results verify the superiority of solving Eqn.~\eqref{eqn:newode} regarding error control.
% Besides, we also compare the FID values by using our improved algorithm with existing fast sampling methods in~\cite{lu2022dpm,zhang2022fast}. 

\subsection{Sampling Quality of AdjointDPM}
\label{subsec:fid}
To evaluate the effectiveness of the reparameterization in AdjointDPM, we generate images by solving the original ODE~\eqref{eqn:ode2} and the reparameterized one~\eqref{eqn:newode} respectively. We also use other state-of-the-art samplers to synthesize images and compare the sampling qualities (measured by FID). We follow the implementation of DPM in~\cite{song2020score} and use the publicly released checkpoints\footnote{\url{https://github.com/yang-song/score_sde}} (trained on the CIFAR10 dataset) to generate images in an unconditional manner. 
We use the \emph{torchdiffeq} package\footnote{\url{https://github.com/rtqichen/torchdiffeq}} and solve the ODE~\eqref{eqn:ode2} and \eqref{eqn:newode} via the Adams--Bashforth numerical solver with order 4.
We choose a suitable NFE number for ODE solvers so that the DPM can generate content with good quality while not taking too much time. We compare the performance of AdjointDPM (solving the reparameterzied ODEs) to the case of solving the original ones under small NFE regions ($\leq 50$).

\begin{wraptable}{l}{0.5\textwidth}
% \vspace{-1em}
 % \begin{table}[h]
\centering
\caption{FID ($\downarrow$) for VPSDE models evaluated on CIFAR10 under small NFE regions.}
\label{tab:fid}
\scalebox{0.85}{
\begin{tabular}{c|cccc}
\toprule 
 NFE & Solving~\eqref{eqn:ode2} & DPM-solver & Solving~\eqref{eqn:newode}  \\ \midrule
 10 & 9.50 & 4.70 & 4.36 \\
 20 & 8.27 & 2.87 & 2.90 \\
 50 & 5.64 & 2.62 & 2.58 \\
 \bottomrule
\end{tabular}
}
\end{wraptable}
We generate the same number of images as the training set and compute the FID between the generated images and the real ones. 
From Table~\ref{tab:fid}, we observe that, after reparameterizing the forward generation process to a non-stiff ODE function, we can generate higher-quality samples with lower FID values under the same NFEs. The sampling qualities of our method are also comparable to those of the state-of-the-art sampler (DPM-solver~\citep{lu2022dpm}).
% Besides, compared with the performance of existing algorithms on CIFAR10, including DPM-solver~\cite{lu2022dpm} and DEIS~\cite{zhang2022fast}, our methods also obtain comparable performance under small NFE regions.

\vspace{-0.5em}
\section{Applications}
\label{sec:exp}
\vspace{-0.5em}
In this section, we apply AdjointDPM to perform several interesting tasks, involving optimizing initial noisy states or model weights for performing guided sampling or customized generation. Due to the space limitation, we provide another application using AdjointDPM for converting visual effects into identification prompt embeddings in Appendix \ref{apdx:inversion}.
The experimental results of all applications demonstrate that our method can effectively back-propagate the loss information on the generated images to the related variables of DPMs.

\subsection{Guided Sampling}
\vspace{-0.5em}

In this section, we use AdjointDPM for guided sampling. The guidance is defined by the loss on the output samples, such as the classification score. We aim to optimize the sampling trajectory, $\{\bx_t\}_{t=T}^{1}$, to make the generated images satisfy certain requirements. 
% In Section \ref{subsec:voc}, we use a fine-grained visual classification (FGVC) model as guidance. The FGVC model can distinguish objects with subtle variations between classes, such as predicting the breeds of animals or the species of plants. Under the guidance of a dog-FGVC model, AdjointDPM can accurately generate dog images of a specific breed. 

\subsubsection{Vocabulary Expansion}
\vspace{-0.5em}

\label{subsec:voc}
The publicly released Stable Diffusion model is pre-trained on a very large-scale dataset (\eg, LAION~\citep{schuhmann2022laion5b}), where the images are captioned at a high level. It can generate diverse general objects. However, when using it for synthesizing a specific kind of object, such as certain breeds of animals or species of plants, we may obtain suboptimal results in the sense that the generated images may not contain subtle characteristics. For example, when generating a picture of the ``Cairn" dog, the Stable Diffusion model can synthesize a dog picture but the shape and the outer coat may mismatch. 
\vspace{-1em}
\begin{figure}[thbp!]
    \centering
    \includegraphics[width=.7\textwidth]{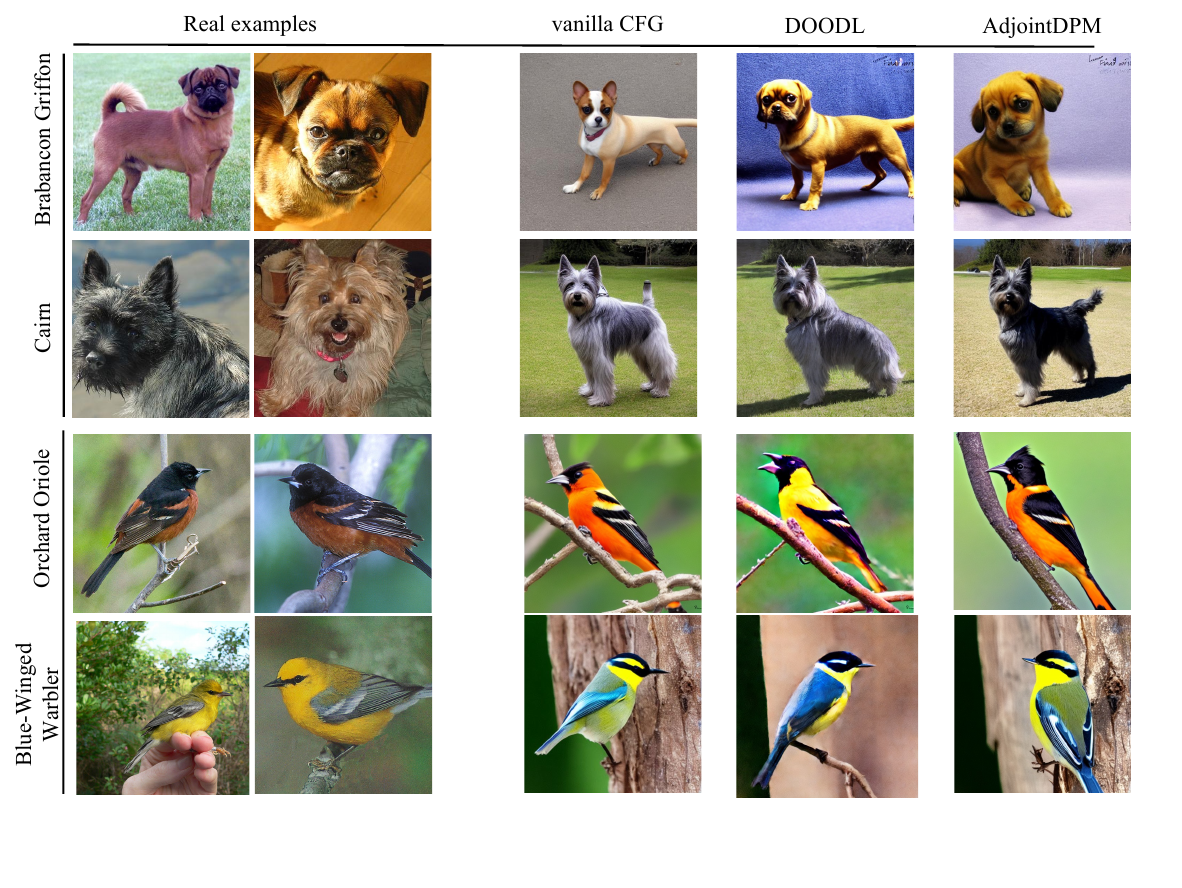}
    \vspace{-2em}
    \caption{Examples for Vocabulary Expansion. The original Stable Diffusion cannot generate samples whose features exactly match the ground-truth reference images. Using the FGVC model, AdjointDPM can guide the Stable Diffusion to synthesize a certain breed of animals. Here we can generate images where the dog's face closely resembles target breeds. Besides, we generate birds with features that are more similar to real images, such as black heads for Orchard Oriole and blue feathers for the Blue-Winged Warbler.}
    \label{fig:app_voc}
\end{figure}

\begin{wraptable}{r}{0.45\textwidth}
% \vspace{-1em}
\centering
\caption{Per-Class $\%$ FID change $(\downarrow)$.}
\label{tab:app_voc}
\scalebox{0.85}{
\begin{tabular}{c|cc}
\toprule 
%  Prompt     & SD    & DOODL & AdjointDPM  \\ \midrule
%  Griffon    & 13.74	   & 13.28     & 13.56 \\
%  Cairn      & 7.34     & 9.39     & 6.54 \\
% xxx   & 13.74	   & 13.28     & 13.56 \\
% xxx     & 7.34     & 9.39     & 6.54 \\
Dataset &  DOODL & AdjointDPM \\ \midrule
Dogs  &  $-5.3\%$ & $-7.1\%$ \\
%CUB & $-3.2\%$ & $-\%$ \\
 \bottomrule
\end{tabular}
}
\end{wraptable}
Here, we use a fine-grained visual classification (FGVC) model as guidance to help the Stable Diffusion model generate specific breeds of dogs. The FGVC model can distinguish objects with subtle variations between classes. Under the guidance of a dog-FGVC model, diffusion models can accurately generate dog images of a specific breed. In other words, \textit{the vocabulary base of the diffusion model gets expanded}.
We formulate this task as follows: Let $f(\cdot)$ denote the FGVC model. The guidance $L(y, f(\bx_0))$ is defined as the prediction score of the generated image $\bx_0$ for class $y$. During sampling, in each time step $t$, we obtain the gradient of guidance $L$ with respect to the noisy state $\bx_t$, namely $\frac{\partial L}{\partial \bx_t}$, to drive the sampling trajectory.

% \begin{wrapfigure}{l}{0.5\textwidth}
%     \centering
%     \includegraphics[width=1\linewidth]{images/app_voc.png} 
%     \caption{Caption1}
%     \label{fig:app_voc}
% \end{wrapfigure}

We present the visual and numerical results in Fig.~\ref{fig:app_voc} and Table~\ref{tab:app_voc},  respectively. By visual comparison to the vanilla Stable Diffusion (SD), we observe that under the guidance of AdjointDPM, the color and outer coat of the generated dog images align better with the ground-truth reference pictures.  Besides, we compute the reduced FID values on 
%Caltech-UCSD Birds (CUB) ~\citep{WahCUB_200_2011}, 
Stanford Dogs (Dogs)~\citep{KhoslaYaoJayadevaprakashFeiFei_FGVC2011} and find that the FID values are also improved. We also compare AdjointDPM to a state-of-the-art baseline, DOODL~\citep{wallace2023endtoend}. AdjointDPM outperforms in terms of visual quality and reduced FID values compared to SD. Refer to more results,  optimization details, and comparison with the existing models in the Appendix~\ref{apdx:vocaexp}.

\subsubsection{Security Auditing}
\label{subsec:security}

% The last application considers using AdjointDPM for auditing the security of AI generation systems. 
DPMs like Stable Diffusion have been widely used in content creation platforms. The large-scale datasets used for training may contain unexpected and harmful data (\eg, violence and pornography). To avoid generating NSFW content, AI generation systems are usually equipped with a safety filter that blocks the outputs of potentially harmful content. However, deep neural networks have been shown to be vulnerable against adversarial examples~\citep{goodfellow_explaining_2015, yan_robustness_2022}.
% including deep classifiers, detection models, \etc 
\textit{This naturally raises the concern---may existing DPM generation systems output harmful content that can sneakily bypass the NSFW filter}? Formally, denote $f(\cdot)$ as the content moderation filter and $c$ as the conditioning prompts containing harmful concepts. We randomly sample an initial noisy state $\bx_T$, the generated image $\Phi(\bx_T, c, \epsilon_{\theta})$ will likely be filtered out by $f(\cdot)$. We want to audit the security of generation systems by searching for another initial noisy state $\bx'_T$, which lies in a small $\delta$-neighborhood of $\bx_T$, such that the corresponding output $\Phi(\bx'_T, c, \epsilon_{\theta})$ may still contain harmful content but bypass $f(\cdot)$. If we find it, the generation systems may face a serious security issue.

This problem can also be formulated as a guided sampling process. The guidance $L$ is defined as the distance between harmful prompt $c$ and the prediction score $f(\Phi(\bx'_T, c, \epsilon_{\theta}))$. The distance is measured by the similarity between CLIP embeddings.
We optimize the perturbation $\delta$ on $\bx_T$ to maximize the distance. The norm of the perturbation is limited to be $\tau$ as we want to ensure the newly generated image is visually similar to the original one.
$$\max_{\delta: \|\delta\|_{\infty} \leq \tau} L(c, f(\Phi(\bx_T+\delta, c, \epsilon_{\theta}))).$$

\begin{figure}[t!]
    \centering
    \includegraphics[width=1\textwidth]{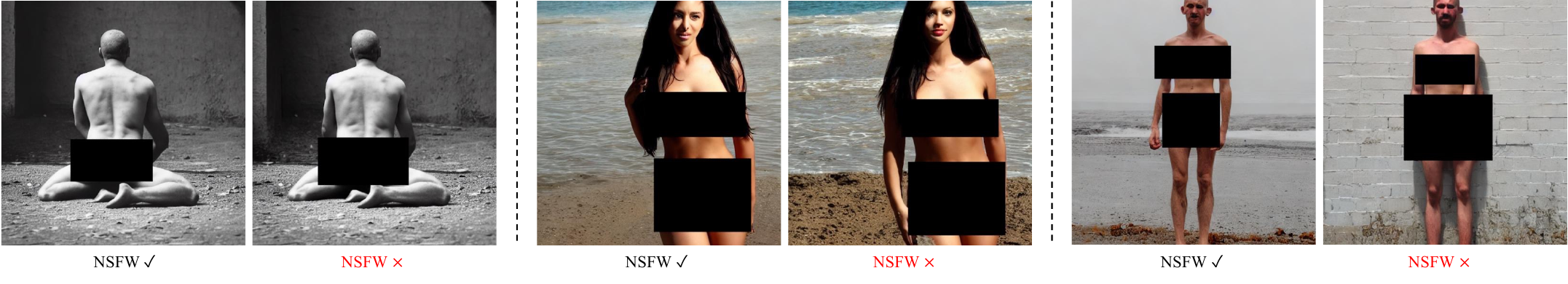}
    % \caption{Left: Adversarial samples against an Imagenet classifier. We show the originally generated images with their class names on the left; these images are correctly classified. On the right, we show the corresponding adversarial images which successfully mislead the classifier. Right: Adversarial samples against the NSFW filter. We show the image generated by conditioning on the prompt \emph{``A photograph of a naked man''} on the left. This image will be blocked by the NSFW filter. However, the right adversarial image circumvents the NSFW filter (Black squares are added by authors for publication).}
    \caption{Adversarial samples against the NSFW filter. We show the image generated by conditioning on harmful prompts (\eg, \emph{``A photograph of a naked man''}) on the left. These images will be blocked by the NSFW filter. However, the images generated from adversarial initial noises circumvent the NSFW filter (Black squares are added by authors for publication).}
    \label{fig:adversary}
    \vspace{-1em}
\end{figure}

We use AdjointDPM to solve this optimization problem and find that there indeed exist initial noisy states of the Stable Diffusion model, whose corresponding outputs can mislead the NSFW filter. Results are shown in Fig.~\ref{fig:adversary}. This observation raises an alert about the security issues of AI generation systems. \textit{Our research community has to develop more advanced mechanisms to ensure the isolation between NSFW content and users, especially teenagers.} 
% More details, including numerical evaluation, are shown in Appendix A.3. 

\begin{wraptable}{l}{0.45\textwidth}
\vspace{-1em}
\centering
\caption{Success ratio (\%) of adversarial initial states bypass the classifier. We show the results for five classes from the ImageNet dataset.}
\label{tab:attack} 
\scalebox{0.85}{
\begin{tabular}{c|ccccc}
\toprule 
Index    & 242    & 430  & 779   & 859 & 895 \\ \midrule
 Ratio          & 63.9	& 75.8  & 45.3 & 58.22 & 52.6 \\
 \bottomrule
\end{tabular}
}
\end{wraptable}
For numerical evaluation, we also audit the security of a smaller diffusion model which is trained on the ImageNet dataset. We chose ten classes and sampled hundreds of images. For each sample, we search for the adversarial perturbation that maximizes the classification error. We record the resultant noisy states that can mislead the classification module. The success ratio achieves around \textit{51.2\%}. Qualitative results about ImageNet are shown in Fig.~\ref{fig:adv_sup} and more experimental details are provided in Appendix~\ref{apdx:security}.

\subsection{Stylization via a Single Reference}
\label{subsec:style}

\begin{figure}[thb!]
    \centering
    \includegraphics[width=\textwidth]{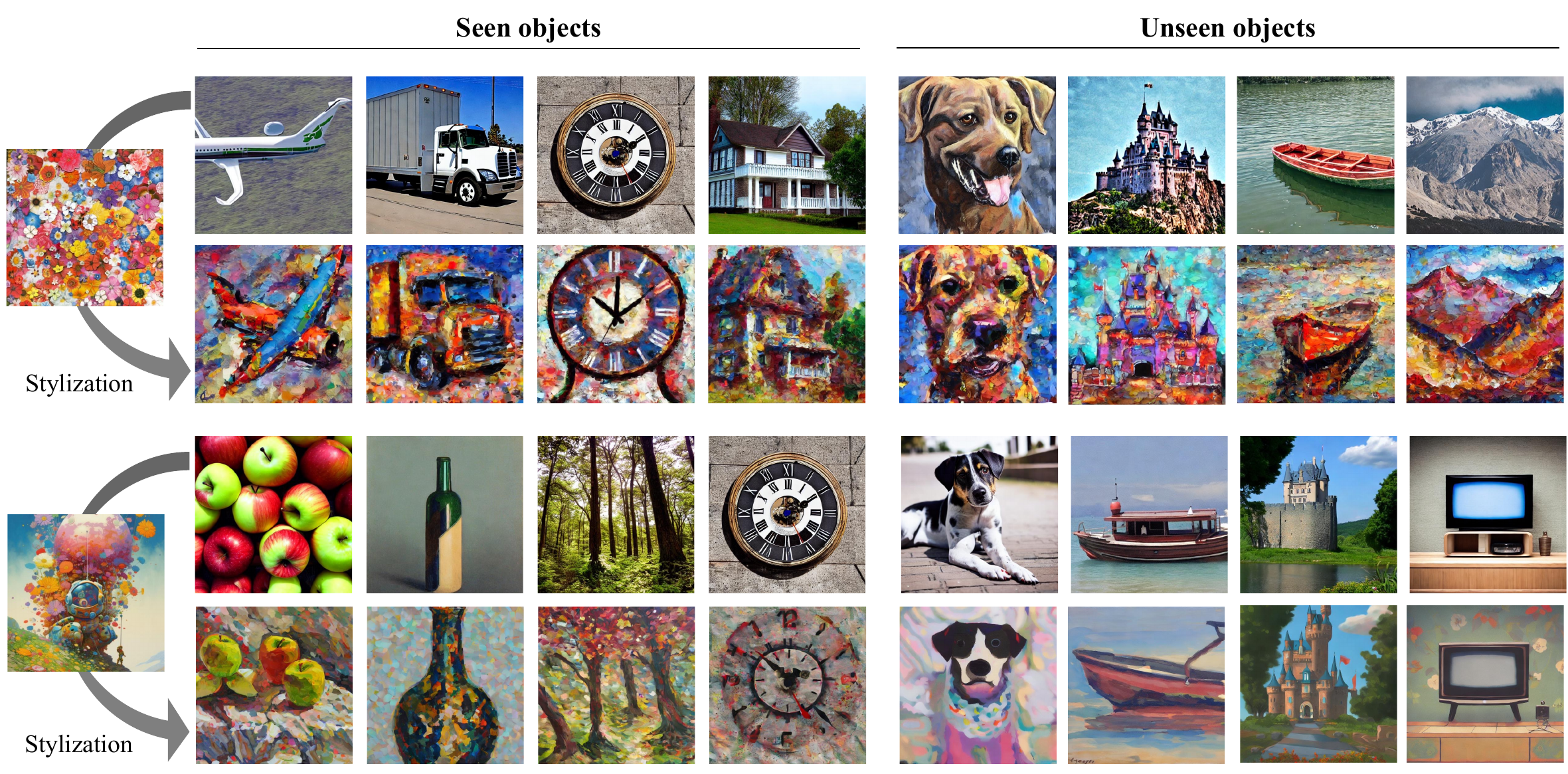}
    \caption{Stylization examples. Images generated by the original Stable Diffusion are shown at the top. The bottom are samples of the stylized Stable Diffusion.}
    \label{fig:style}
\end{figure}

We consider using AdjointDPM to fine-tune the weights of the UNet in Stable Diffusion for stylization based on a single reference image. We define the style of an image $\bx$ by its Gram matrix~\citep{gatys_neural_2015}, which is computed based on the features extracted from a pre-trained VGG model.\footnote{\url{https://pytorch.org/tutorials/advanced/neural_style_tutorial.html}} Here we denote the features extracted from the VGG as $\mathbf{F}(\bx)$ and the Gram matrix $\mathbf{G}(\bx)=\bF\bF^T$. 

Given a reference image, we denote the target style as $\bG_{\mathrm{style}}$. We aim to fine-tune the weights of the UNet so that the style of generated images matches the target one. This task is formulated as the optimization problem~\eqref{task:stylization}. The objective contains two terms. Besides the style loss $ L_{\mathrm{s}}$ (mean squared error), we also add a term of content loss $ L_{\mathrm{c}}$. The content loss encourages the model to preserve the ability to generate diverse objects when adapting itself to a certain style. In specific, we sample multiple noise-prompt-image triplets, $\{(\bx_T^i, c^i, \bx_0^i)\}_{i=1}^{N}$, where $\bx_0^i$ denotes the clean image generated by the pre-trained Stable Diffusion with the input of $(\bx_T^i, c^i)$. The content loss is defined as the mean squared error between the features of originally generated images $\bF(\bx_0^i)$ and those generated by optimized weights $\bF(\Phi(\bx_T^i, c^i, \epsilon_{\theta}))$. Two coefficients, $w_{\mathrm{s}}$ and $w_{\mathrm{c}}$, balance the strength of the two loss terms. 
\begin{align}
\label{task:stylization}
\min_{\theta}\frac{1}{N}\sum_{i} \left[w_{\mathrm{s}} L_{\mathrm{s}}\left( \bG_{\mathrm{style}}, \bG(\Phi(\bx_T^i, c^i, \epsilon_{\theta})) \right) + w_{\mathrm{c}} L_{\mathrm{c}}\left( \bF(\bx^i), \bF(\Phi(\bx_T^i, c^i, \epsilon_{\theta})) \right)\right]
\end{align}

\begin{wraptable}{l}{0.5\textwidth}
% \vspace{-1em}
 % \begin{table}[h]
\centering
\caption{CLIP similarity scores ($\uparrow$) between samples and the conditioning prompts.}
\label{tab:stylization}
\scalebox{0.85}{
\begin{tabular}{c|ccc}
\toprule 
 Prompt & DreamBooth & Text-Inv & AdjointDPM  \\ \midrule
 % Average    & 26.03	   & 24.40     & 28.35 \\
 Airplane   & 21.98     & 24.44     & 27.34 \\
 Clock      & 28.23     & 26.74     & 30.00 \\
 House      & 25.40     & 25.91     & 29.03 \\
 Cat        & 25.90     & 22.30     & 26.83 \\
 Apples     & 28.10     & 24.06     & 28.59 \\
 \bottomrule
\end{tabular}
}
% \vspace{-1em}
% \end{table}
\end{wraptable}
We construct 10 prompts corresponding to ten of the CIFAR100 classes and sample starting noises to generate 100 images (10 images for each prompt) to form our training dataset. Visual and numerical results are shown in Fig.~\ref{fig:style} and Table~\ref{tab:stylization} respectively. We observe the SD model fine-tuned by AdjointDPM can generate stylized images of different objects. The stylizing capability also generalizes to the concepts unseen during fine-tuning (shown in the right part of Fig.~\ref{fig:style}). In addition to the high visual quality, the samples also align well with the conditioning prompts according to the high CLIP similarity scores. We compare AdjointDPM with other methods for stylization, including DreamBooth~\citep{ruiz_dreambooth_2022} and Textural-Inversion~\citep{gal_image_2022} (see the qualitative comparisons in Fig.~\ref{fig:text_inver}). We observe that AdjointDPM achieves better alignment between image samples and the prompts. In addition, these existing methods barely can be generalized to unseen objects in this case only one reference image is available. More details and examples of stylization are shown in Appendix~\ref{apdx:style}. 

% \vspace{-0.5em}

\section{Related Works and Discussion}
\paragraph{Customization of Text-to-Image Generation}
Text-to-image customization aims to personalize a generative model for synthesizing new images of a specific target property. Existing customization methods typically tackle this task by either representing the property via a text embedding~\citep{gal_image_2022,mokady_null-text_2022,wen_hard_2023} or finetuning the weights of the generative model~\citep{ruiz_dreambooth_2022, kawar_imagic_2023, han_svdiff_2023}.
%%%% The following sentences are taken from other paper. Need to rewrite!
For example, Textual-Inversion~\citep{gal_image_2022} inverts the common identity shared by several images into a unique textual embedding. To make the learned embedding more expressive, \citet{daras_multiresolution_2022} and \citet{voynov_p_2023} generalize the unique embedding to depend on the diffusion time or the layer index of the denoising UNet, respectively. In the other line, DreamBooth~\citep{ruiz_dreambooth_2022} learns a unique identifier placeholder and finetunes the whole diffusion model for identity customization. 
To speed up and alleviate the overfitting, Custom Diffusion~\citep{kumari_multi-concept_2022} and SVDiff~\citep{han_svdiff_2023} only update a small subset of weights. 
Most of these existing methods assume that a handful of image examples (at least 3-5 examples) sharing the same concept or property are provided by the user in the first place. Otherwise, the generalization of resultant customized models usually will be degraded, \ie, they barely can synthesize unseen objects (unseen in the training examples) of the target concept. 
% Concerning the customization applications, most existing works \citep{gal_image_2022, wen_hard_2023, kumari_multi-concept_2022, ruiz_dreambooth_2022} require collecting samples that share a common property (e.g., the same identity, style, or visual effect), then finetune the network weights or text prompts on these reference sample. 
% To ensure the learned parameters can represent the target property and generalize well, these works usually require collecting enough samples or designing dedicated regularization tricks. 
However, in some cases, it is difficult or even not possible to collect enough data that can represent abstract requirements imposed on the generated content. For example, we want to distill the editing effects/operations shown in a single image by a media professional or a novel painting style of a unique picture. In contrast, this paper relaxes the requirement of data samples and proposes the AdjointDPM for model customization only under the supervision of a differentiable loss.

\paragraph{Guidance of Text-to-Image Generation} Concerning the guidance of diffusion models, some algorithms~\citep{bansal_universal_2023, yu2023freedom} mainly use the estimated clean state for computing the guidance loss and the gradient with respect to intermediate noisy states. The gap to the actual gradient is not negligible. As a result, the guided synthesized results may suffer from degraded image quality. Instead, some other methods~\citep{wallace2023endtoend, Liu_2023_CVPR}, such as DOODL, compute the gradients by exploiting the invertibility of diffusion solvers (\eg, DDIM). This paper also formulates the sampling process as ODE and proposes AdjointDPM to compute the gradients in a more accurate and adaptive way. 

The main contribution of this paper is that we propose a ubiquitous framework to optimize the related variables (including UNet weights, text prompts, and latent noises) of diffusion models based on the supervision information (any arbitrary differentiable metric) on the final generated data. To our best knowledge, AdjontDPM is the first method that can compute the gradients for all types of parameters of diffusion models. In contrast, other methods, such as DOODL~\citep{wallace2023endtoend}, FlowGrad~\citep{Liu_2023_CVPR}, and DEQ-DDIM~\citep{pokle2022deep}), either only work for the noisy states or require the diffusion sampling process having equilibrium points. In the future, we will explore more real-world applications of AdjointDPM.

\subsubsection*{Acknowledgments}
This research/project is supported by the Singapore Ministry of Education Academic Research Fund (AcRF) Tier 2 under grant number A-8000423-00-00 and the Singapore Ministry of Education AcRF Tier 1 under grant number A-8000189-01-00. We would like to acknowledge that the computational work involved in this research work is partially supported by NUS IT’s Research Computing group. 

%\newpage
\bibliography{iclr2024_conference}
\bibliographystyle{iclr2024_conference}

\newpage
\appendix
\tableofcontents

% \newpage
\section{Text Embedding Inversion}
\label{apdx:inversion}
In addition to the applications shown in Section~\ref{sec:exp}, we consider another application concerning using AdjointDPM to convert visual effects (\eg, bokeh and relighting) into an identification text embedding \cinvtoken.
Suppose we are given an image pair, namely an original image and its enhanced, the enhanced version is edited by some professional and appears with certain fascinating visual effects. After optimization, we can combine the obtained embedding with various text prompts to generate images with the same visual effect. Here, we simulate this real setting by using a text-to-image model to generate images with and without a certain effect. 

Suppose we are provided a text-to-image DPM $\Phi(\cdot)$, we can generate an image $\bx$ by denoising randomly sampled noise $\bx_T$ in the condition of the base prompt $c_{\text{base}}$. 
% The generated image $\bx$ can be further edited by a professional photographer to improve the aesthetic quality (\eg, color, tone, and bokeh). We want to distill the editing effect in the edited image $\bx^*$ into a special embedding \invtoken. 
We further improve the aesthetic quality by inserting some keywords $c_{\text{target}}$, like ``bokeh'', into the conditioning prompt. The newly generated images are denoted by $\bx^*$. 
We use $\bx^*$ or its feature as a reference to define a loss $L(\cdot)$ that measures the distance to the target effect, such as the $\ell_2$ or perceptual loss. We aim to optimize a special embedding \invtoken ~that can recover the visual effects in $\bx^*$:
% ~can be obtained by solving the following optimization problem:
\begin{align*}
\min_{\text{\cinvtoken}}L\left( \bx^*, \Phi(\bx_T, \{c_{\text{base}}, \text{\invtoken}\}, \epsilon_{\theta}) \right).
\end{align*}

\begin{figure}[htbp]
    \centering
    \includegraphics[width=\textwidth]{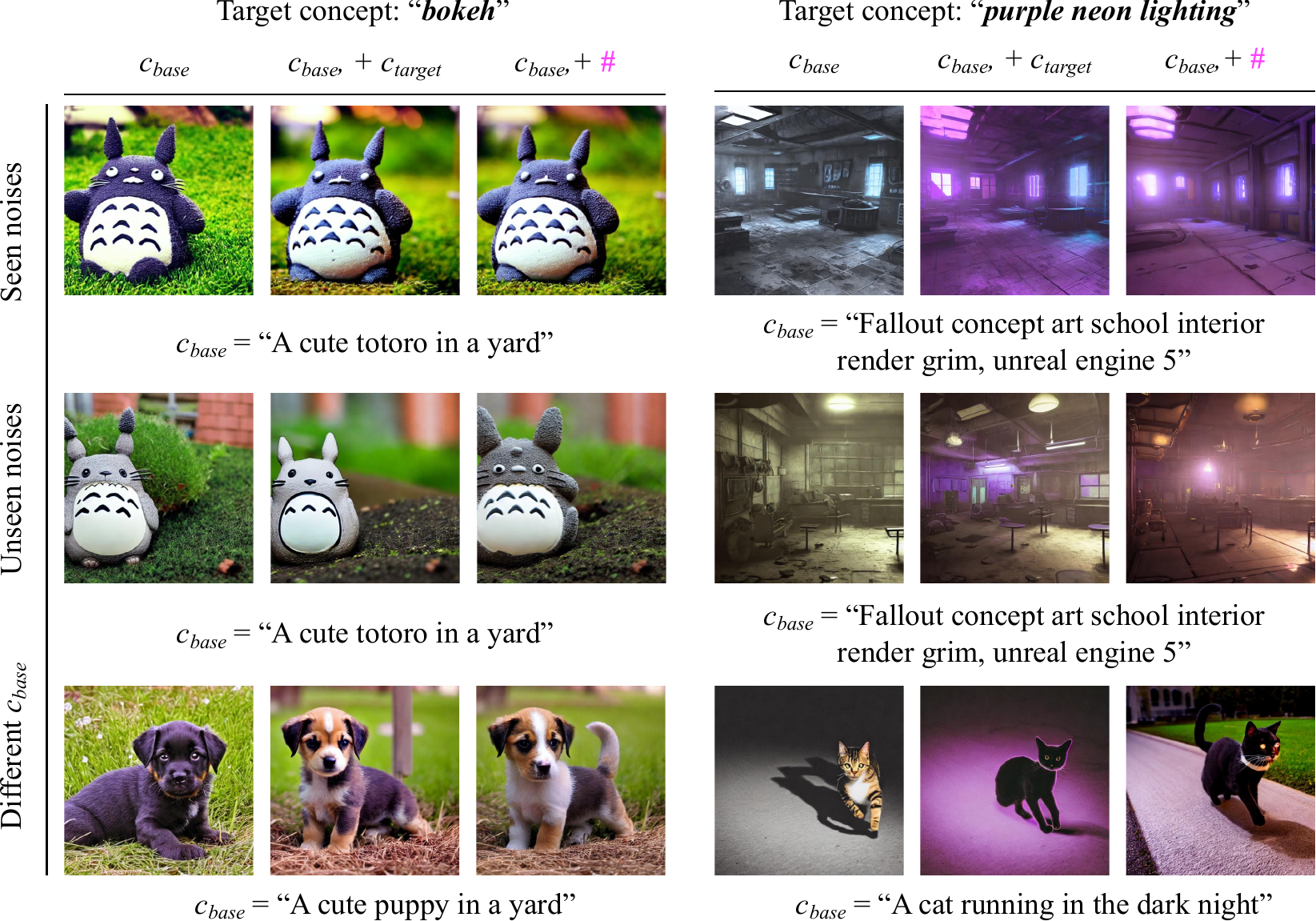}
    \caption{Examples on prompt inversion - part 1}
    \label{fig:text_embed}
\end{figure}

We utilize the publicly released Stable Diffusion models for image generation and set the loss function as the mean squared error (MSE) between the target images and the generated images. We aim to optimize a prompt embedding in the CLIP~\citep{radford_learning_2021} embedding space and use the obtained embedding for image generation by concatenating it with the embeddings of other text prompts. 
% We first generate an image with a text prompt $c$ and randomly sampled noise $\bx_T$. To improve the visual quality of the generated image,  we either manually edit it by adjusting the color and tone or concatenate the prompt $c$ with an effect token $c^*$ and generate another image $\bx^*$ on the condition of $\{c, c^*\}$.
% We optimize the special embedding \invtoken ~to recover the editing effect.
As shown in Fig.~\ref{fig:text_embed}, we observe AdjointDPM successfully yields an embedding \invtoken ~that can ensure the appearance of target visual effects, including the \textit{bokeh} and \textit{relighting}. Furthermore, the obtained embeddings \invtoken ~also generalize well to other starting noise and other text prompts. For example, the bokeh-\invtoken~is optimized on a pair of \textit{totoro} images; it also can be used for generating different images of \textit{totoro} and other objects like \textit{dog}. Similarly, the obtained \invtoken~corresponding to manual editing (``converting to black and white'') also can be used for novel scene generation. 
% More examples are shown in Appendix A.1. 

% \section{More Examples on Text Embedding Inversion}
% % \subsection{More Examples on Text Embedding Inversion}
% \label{app:text}

% To optimize a suitable text embedding for unknown concepts, we fix the text embedding of $c_{\mathrm{base}}$ and fine-tune the remaining text embedding for unknown effects. Then we minimize the MSE loss between initially generated images and target images. We run 50 epochs with AdamW optimizer of learning rate $0.1$.  In addition to visual effects \emph{bokeh} and \emph{relighting} we show in Sec. 4.1, we can also obtain text embeddings for other visual effects such as 
% %\emph{black and white} and 
% \emph{stylization}. We present more examples of text embedding inversion in Fig.~\ref{fig:text2}.  

\begin{figure}[htbp!]
    \centering
    \includegraphics[width=\textwidth]{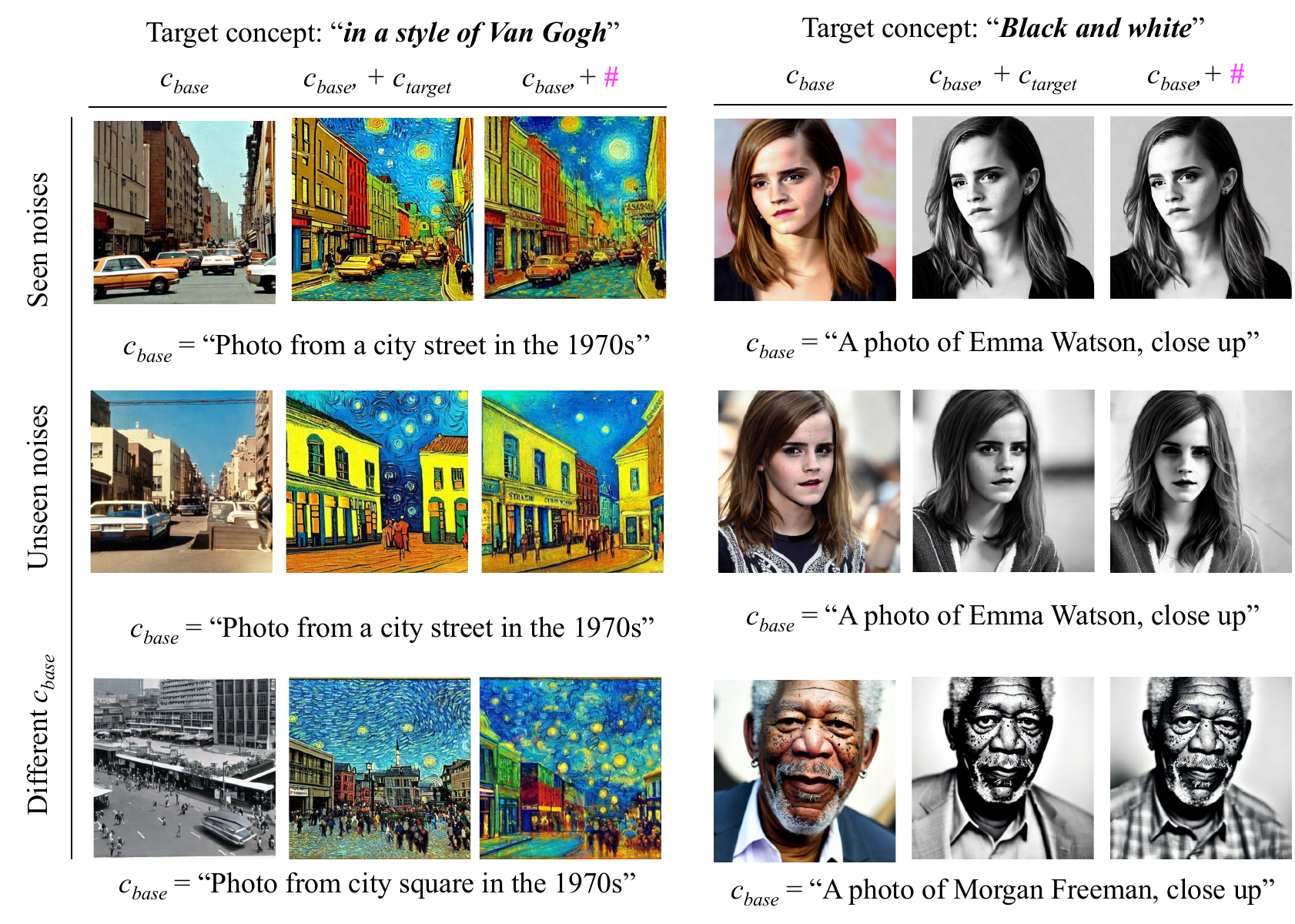}
    \caption{Examples on prompt inversion - part 2}
    \label{fig:text2}
\end{figure}

\newpage
\section{Experimental Details on Vocabulary Expansion and Comparison with Existing Models}
\label{apdx:vocaexp}

\paragraph{Comparison with Existing Models.} 
    DOODL~\citep{wallace2023endtoend} optimizes the initial diffusion noise vectors w.r.t a model-based loss on images generated from the full-chain diffusion process. In their work, they obtain the gradients of loss w.r.t noise vectors by using invertible neural networks (INNs). There are three main differences between DOODL and our work. First, while DOODL optimizes the initial diffusion noise vectors, our work optimizes related variables, including network parameters, initial noises and textual embeddings w.r.t a model-based loss on images generated from the full-chain diffusion process. Thus, we consider the broader cases of DOODL. Second,  in the calculation of gradients w.r.t initial noises, DOODL uses the invertibility of EDICT~\citep{wallace2023edict}, i.e., $x_0$ and $x_T$ are invertible. This method does not apply to the calculation of gradients w.r.t. network parameters and textual embeddings as they share across the full-chain diffusion process. Finally, with regard to the memory consumption when calculating gradients with respect to the initial noise, our experimental results are as follows: we utilized the stable diffusion v1.5 checkpoint to run both the AdjointDPM and DOODL models on a V100 GPU (32GB memory). For the AdjointDPM method, backpropagating the gradients with respect to a single initial noise required 19.63GB of memory. In comparison, the DOODL method consumed 23.24GB for the same operation.  The additional memory consumption in DOODL is mainly from the dual diffusion process in EDICT. Thus, our method is more efficient in memory consumption. In terms of time consumption, DOODL relies on the invertibility of EDICT, resulting in identical computation steps for both the backward gradient calculation and the forward sampling process. Besides, they usually use DDIM sampling methods, which is equivalent to the first-order neural ODE solver. However, our AdjointDPM methods have the flexibility to apply high-order ODE solvers, allowing for faster backward gradient calculation. See the following for an experimental comparison between our method and DOODL. 
    
    We also make a comparison with FlowGrad~\citep{Liu_2023_CVPR}. FlowGrad efficiently backpropagates the output to any intermediate time steps on the ODE trajectory, by decomposing the backpropagation and computing vector Jacobian products. FlowGrad focuses on refining the ODE generation paths to the desired direction. This is different from our work, which focuses on the finetuning of related variables, including network parameters, textual embedding and initial noises of diffusion models for customization. Besides, FlowGrad methods also can not obtain the gradients of loss w.r.t. textual embeddings and neural variables as these variables share across the whole generation path. Then for the gradients w.r.t the latent variables, we could show the memory consumption of our methods is constant while they need to store the intermediate results. 

\paragraph{Experimental Details on Vocabulary Expansion}

\begin{figure}[thbp!]
    \centering
    \begin{subfigure}{\textwidth}
    \includegraphics[width=\linewidth]{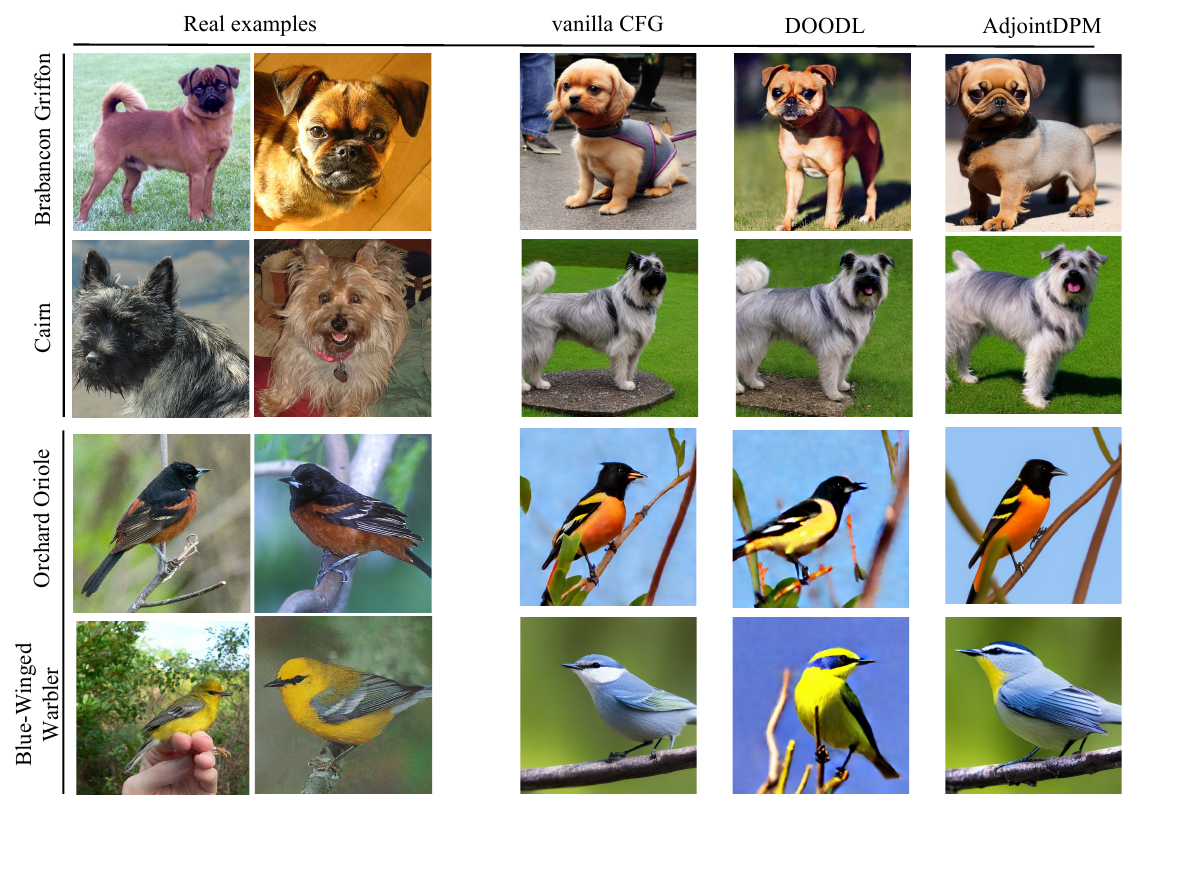}
    %\caption{More Examples for Vocabulary Expansion. }
    %\label{fig:voc_app}
    \end{subfigure}
    
    \begin{subfigure}{\textwidth}
    \includegraphics[width=\linewidth]{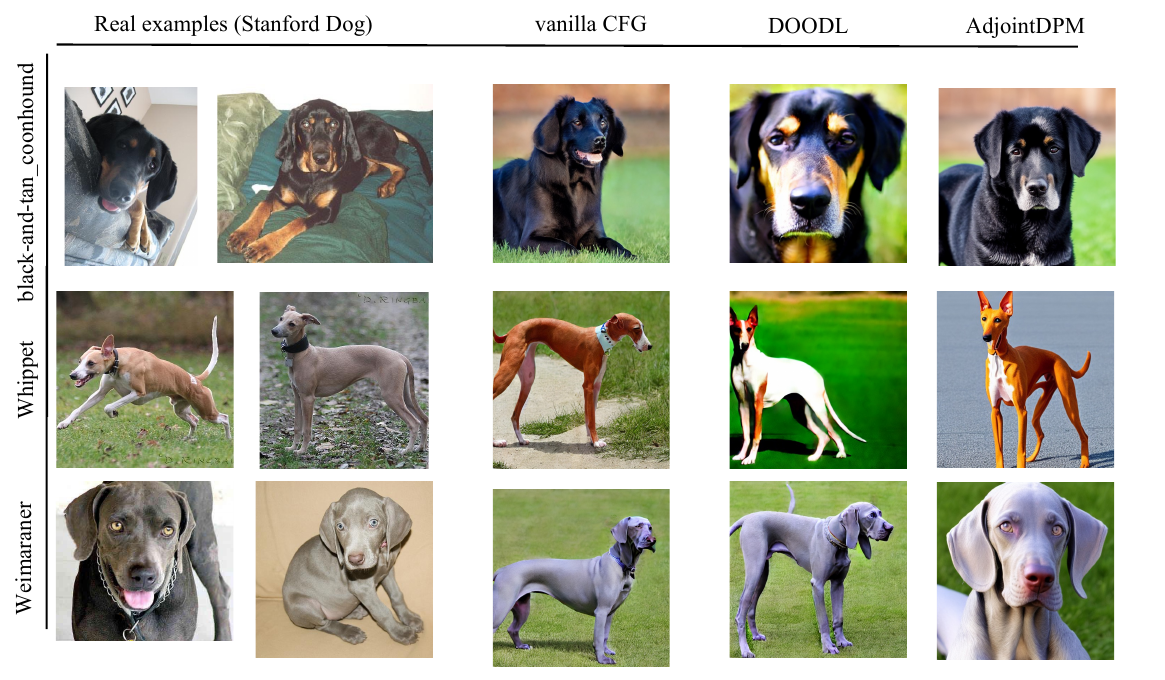}
    %\caption{More Examples for Vocabulary Expansion. }
    %\label{fig:voc_app}
    \end{subfigure}
   \caption{More Examples for Vocabulary Expansion. }
    \label{fig:voc_app}
\end{figure}

During the optimization of the noise states under the guidance of FGVC model, we adopt the Euler ODE solver in our AdjointDPM method with 31 steps. We optimize the noise states using the AdamW optimizer for 30 epochs with different learning rates for different breeds. For the implementation of DOODL, we follow the officially released code\footnote{\url{https://github.com/salesforce/DOODL}} and we set the sampling steps also to be 31 and optimization steps for 30. Following the DOODL, we also measure the performance by computing the FID between a set of generated images (4 seeds) and the validation set of the FGVC dataset being studied. We do experiments on Stanford Dogs (Dogs)~\citep{KhoslaYaoJayadevaprakashFeiFei_FGVC2011} datasets and calculate FID values. More qualitative results are shown in Fig~\ref{fig:voc_app}.

\section{Experimental Details and More Results on Security Auditing}
\label{apdx:security}

In this section, we provide explicit details about generating adversarial examples against an ImageNet classifier and the NSFW filter in Stable Diffusion, respectively.

\paragraph{Security Auditing under an ImageNet Classifier.}
To generate adversarial samples against an ImageNet classifier, we follow the implementation of classifier guidance generation of DPM\footnote{\url{https://github.com/LuChengTHU/dpm-solver/tree/main/examples/ddpm_and_guided-diffusion}} and use the publicly released checkpoints trained on the ImageNet 128x128 dataset to generate images in a conditional manner. We adopt the pre-trained ResNet50\footnote{\url{https://pytorch.org/vision/stable/models.html}} as our ImageNet classifier. 

\begin{figure}[htbp!]
    \centering
    \includegraphics[width=.6\textwidth]{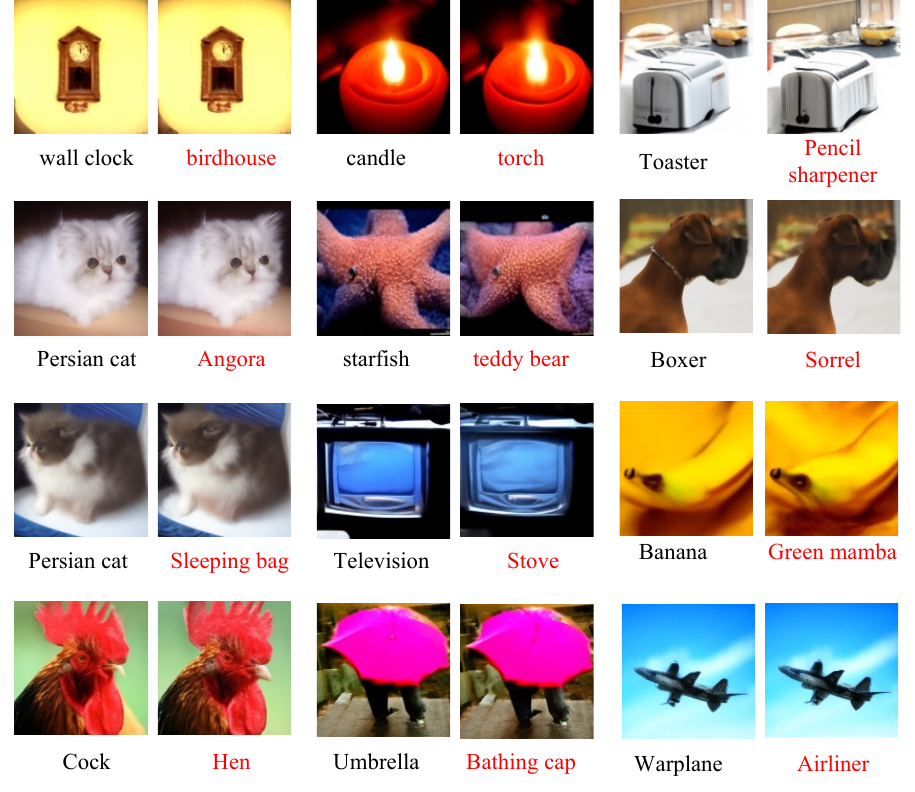}
    \caption{Adversarial examples against the ImageNet classifier. We show the originally generated images with their class names on the left; these images are correctly classified by ResNet50. On the right, we show the corresponding adversarial images which successfully mislead the classifier.}
    \label{fig:adv_sup}
\end{figure}

To generate adversarial examples, we first randomly choose an ImageNet class and set it as the class label for classifier guidance generation, and then we pass the generated images to the ResNet50 classifier. If the outputs of ResNet50 classifier are aligned with the chosen class label, we begin to do an adversarial attack by using AdjointDPM. For the adversarial attack, we adopt the targeted attack, where we choose a target class and make the outputs of ResNet50 close to the pre-chosen target class by minimizing the cross entropy loss. We also clamp the updated initial noise in the range of $[x_T-0.8, x_T+0.8]$ (i.e., set $\tau = 0.8$ in Sec. 4.3 to ensure that generated images do not visually change too much compared with the start images). We show more adversarial examples against the ImageNet classifier in Fig.~\ref{fig:adv_sup}. Besides, define the attack rate as the ratio between the number of samples with incorrect classification results after the attack and the total number of samples. We also get the attack rate $51.2\%$ by generating 830 samples from 10 randomly chosen classes. The class labels here we choose are $[879, 954, 430, 130, 144, 242, 760, 779, 859, 997]$.

\paragraph{Experimental Details on the NSFW Filter.} In this case, we set $\tau=0.9$. We follow the implementation of Stable Diffusion\footnote{\url{https://github.com/huggingface/diffusers}} and set the loss function as the cosine distance between CLIP embeddings~\cite{radford_learning_2021} of generated images and unsafe embeddings from~\cite{rando2022redteaming}.

%\newpage
\section{More Examples on Finetuning Weights for Stylization}
% \subsection{More Examples on Finetuning Weights for Stylization}
\label{apdx:style}

In this section, we introduce the experimental details of stylization and present more stylized examples on seen noises and seen classes, unseen noises and seen classes, and unseen noises and unseen classes. 

\begin{figure}[thbp!]
    \centering
    \includegraphics[width=\textwidth]{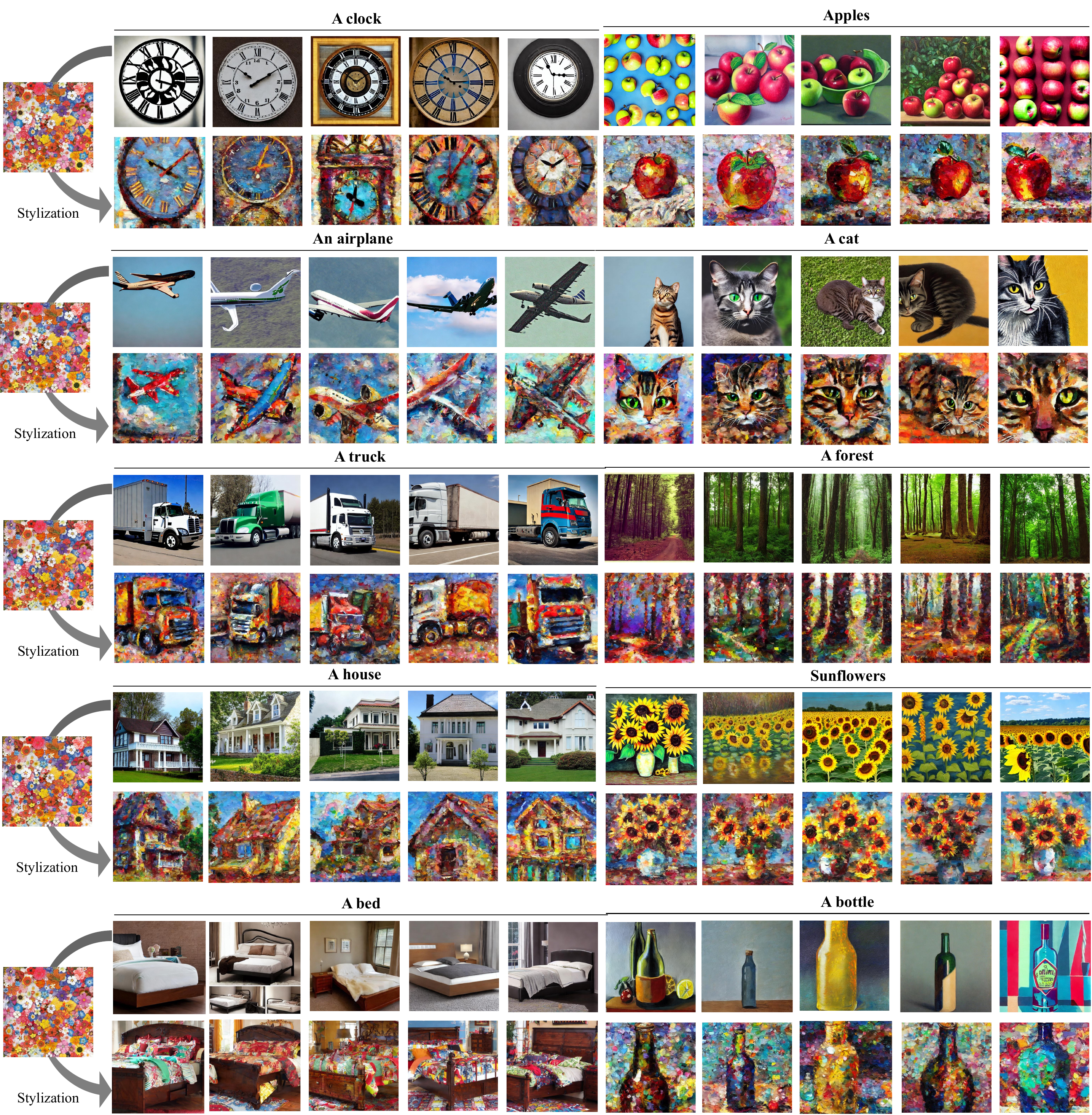}
    \caption{Stylization examples on seen classes and seen noises}
    \label{fig:style_full}
\end{figure}

\begin{figure}[thbp!]
    \centering
    \includegraphics[width=\textwidth]{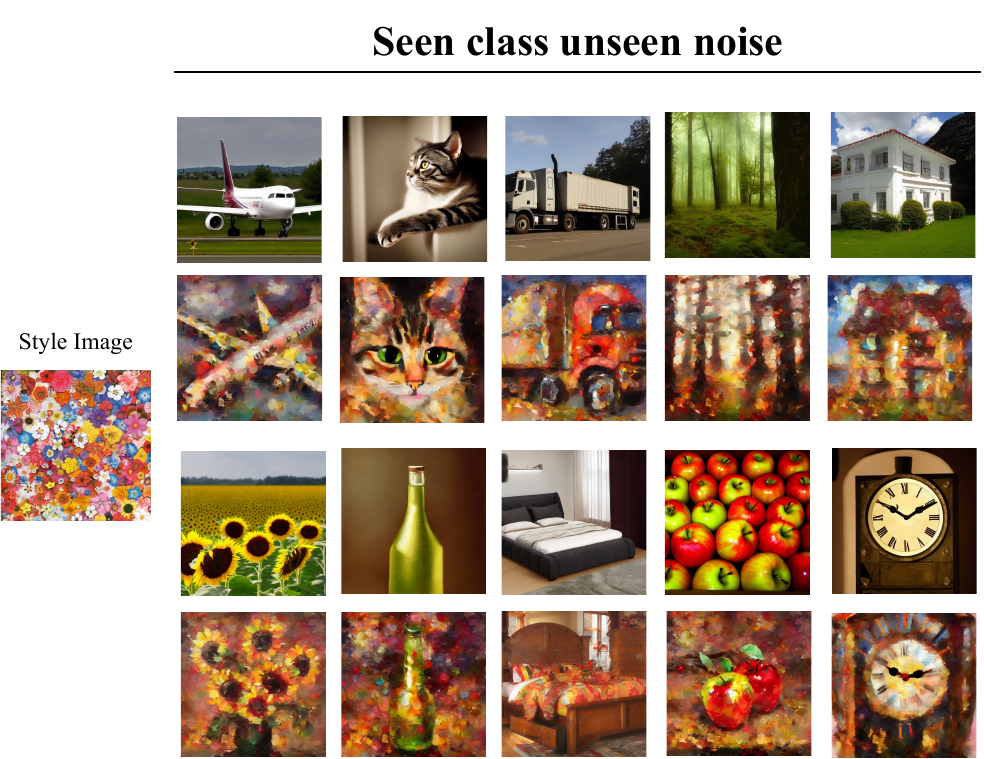}
    \caption{Stylization examples on seen classes and unseen noises using DDPM.}
    \label{fig:app_sde}
\end{figure}

\begin{figure}[thbp!]
    \centering
    \includegraphics[width=1\textwidth]{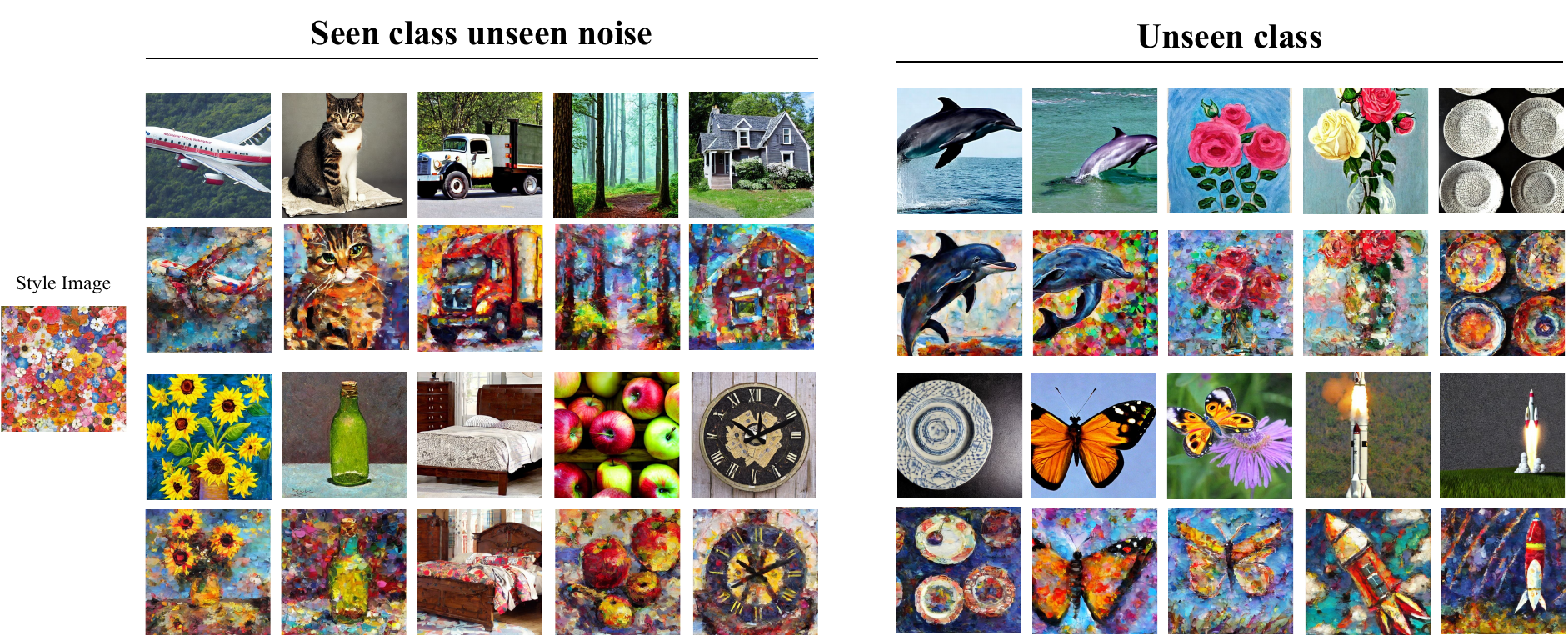}
    \caption{Stylization examples on \textit{seen}-class-\textit{unseen}-noise and \textit{unseen}-class}
    \label{fig:style_gen2}
\end{figure}

\begin{figure}[thbp!]
    \centering
    \begin{subfigure}{\textwidth}
    \includegraphics[width=\textwidth]{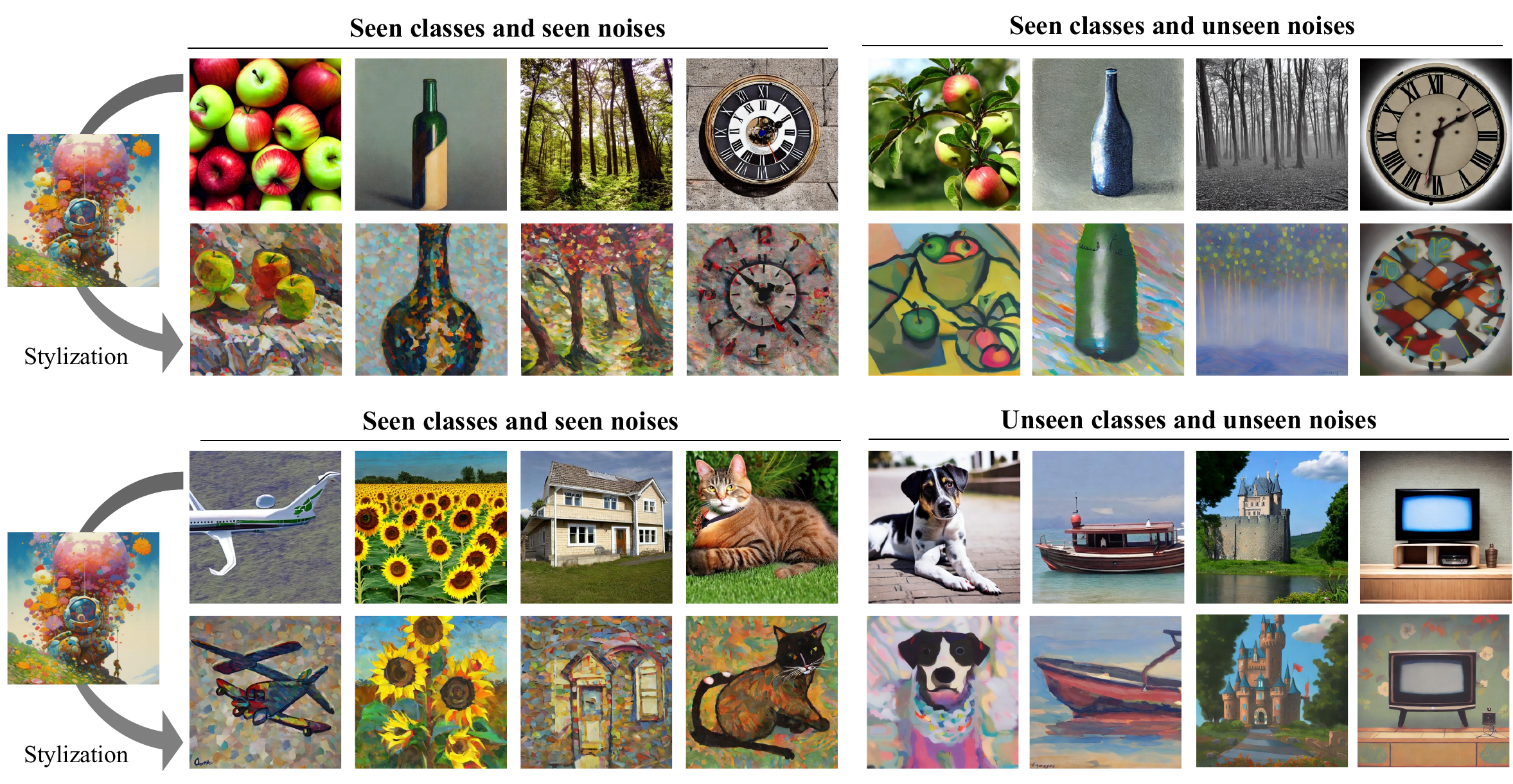}
    \end{subfigure}  
    \begin{subfigure}{\textwidth}
    \includegraphics[width=\textwidth]{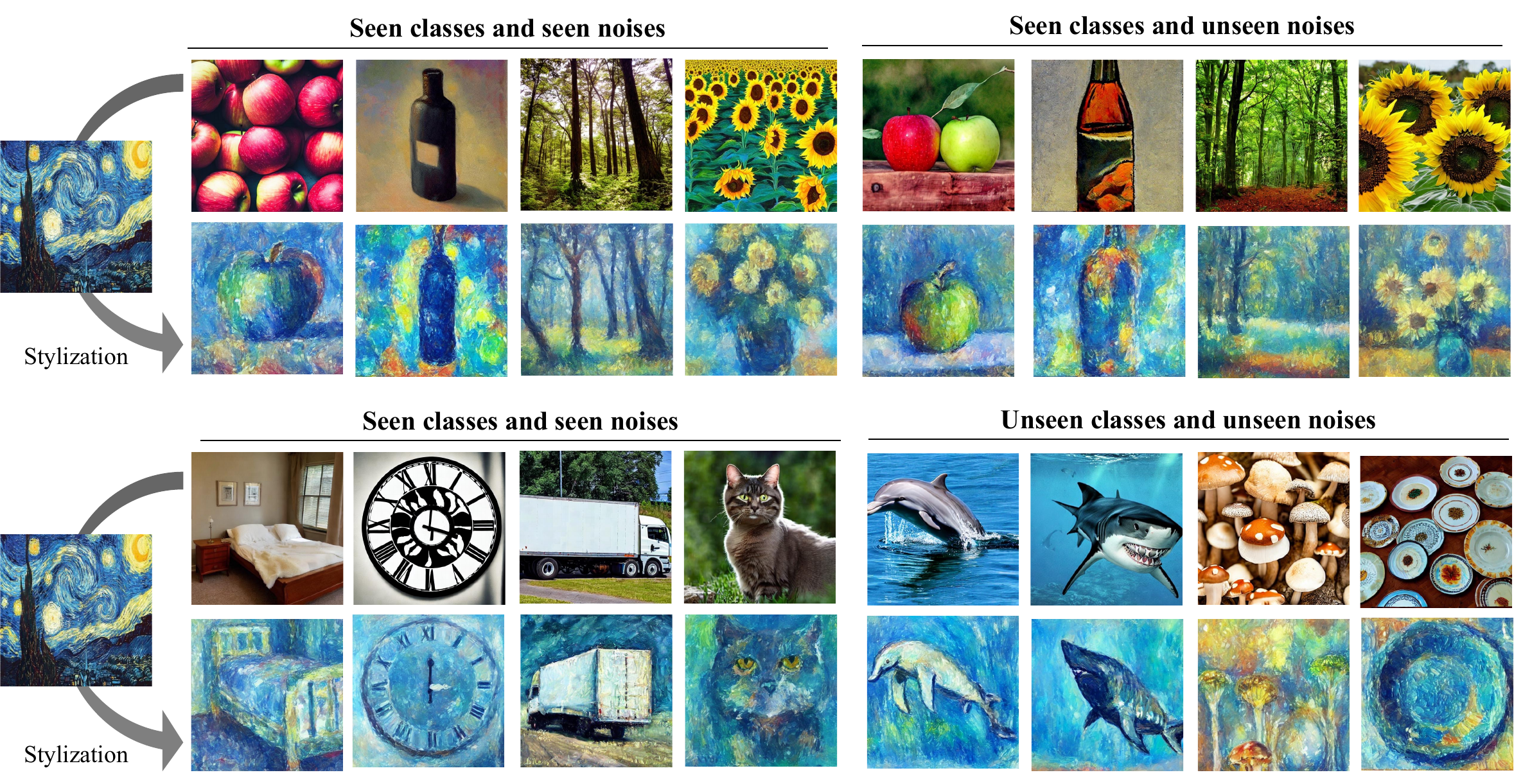}
    \end{subfigure} 
    \caption{Stylization examples on other style images}
    \label{fig:style_mid}
\end{figure}

For training, we choose ten classes from CIFAR-100 classes, which are [``An airplane'', ``A cat'', ``A truck'', ``A forest'', ``A house'', ``sunflowers'', ``A bottle'', ``A bed'', ``Apples'', ``A clock'']. Then we randomly generate 10 samples from each class to compose our training dataset. Besides, we directly use these class names as the input prompt to Stable Diffusion~\footnote{\url{https://github.com/CompVis/stable-diffusion}}. We optimize the parameters of cross attention layers of UNet for 8 epochs by using AdamW optimizer with learning rate $10^{-4}$. We show more stylization results on 100 training samples (seen noises and seen classes) in Fig.~\ref{fig:style_full}. Meanwhile, we also show more examples of seen-classe-unseen-noise and examples of unseen-class-unseen-noise in Fig.~\ref{fig:style_gen2}. In Fig.~\ref{fig:style_mid}, we also show the stylization results on other target style images, in which one is downloaded from the showcase set of Midjourney\footnote{\url{https://cdn.midjourney.com/61b8bd5d-846b-4f69-bdc1-0ae2a2abcce8/grid_0.webp}} and the other is the Starry Night by Van Gogh.  We also present that the finetuned networks under ODE forms can still apply SDE solvers (such as DDPM) in Fig.~\ref{fig:app_sde}.

% \begin{figure}[thbp!]
%     \centering
%     \includegraphics[width=.6\textwidth]{images/unseen_noise.pdf}
%     \caption{Stylization examples on seen classes and \textit{unseen} noises}
%     \label{fig:style_gen1}
% \end{figure}

% \begin{figure}[thbp!]
%     \centering
%     \includegraphics[width=.6\textwidth]{images/unseen_classes.pdf}
%     \caption{Stylization examples on \textit{unseen} classes and \textit{unseen} noises}
%     \label{fig:style_gen2}
% \end{figure}

\subsection{Qualitative comparisons to Textual-Inversion and DreamBooth}
\begin{figure}[thbp!]
    \centering
    \begin{subfigure}{.65\textwidth}
    \includegraphics[width=\textwidth]{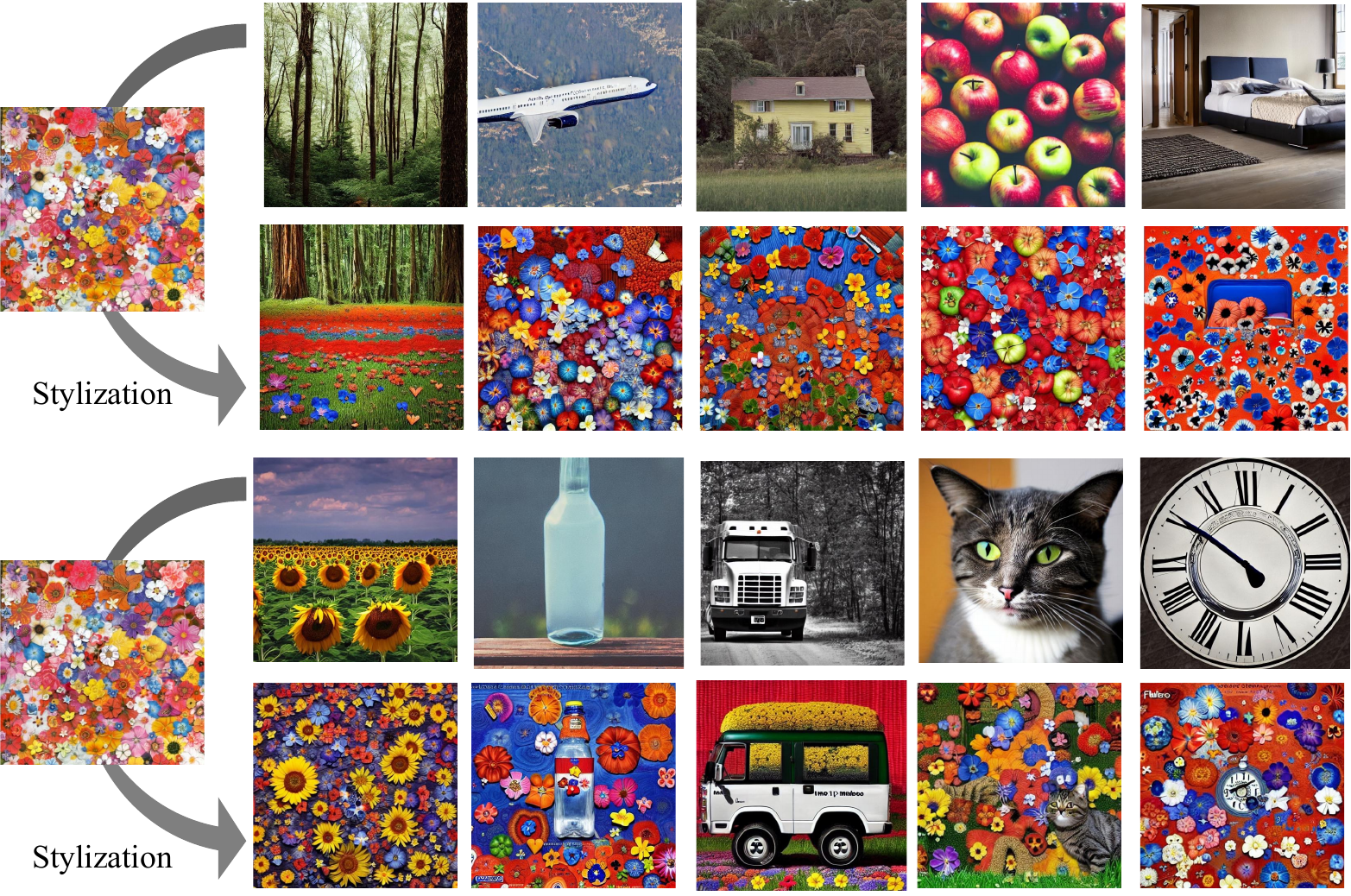}
    \caption{Stylization examples generated by Textual Inversion.}
    \end{subfigure}
    ~\\
    \begin{subfigure}{.65\textwidth}
    \includegraphics[width=\textwidth]{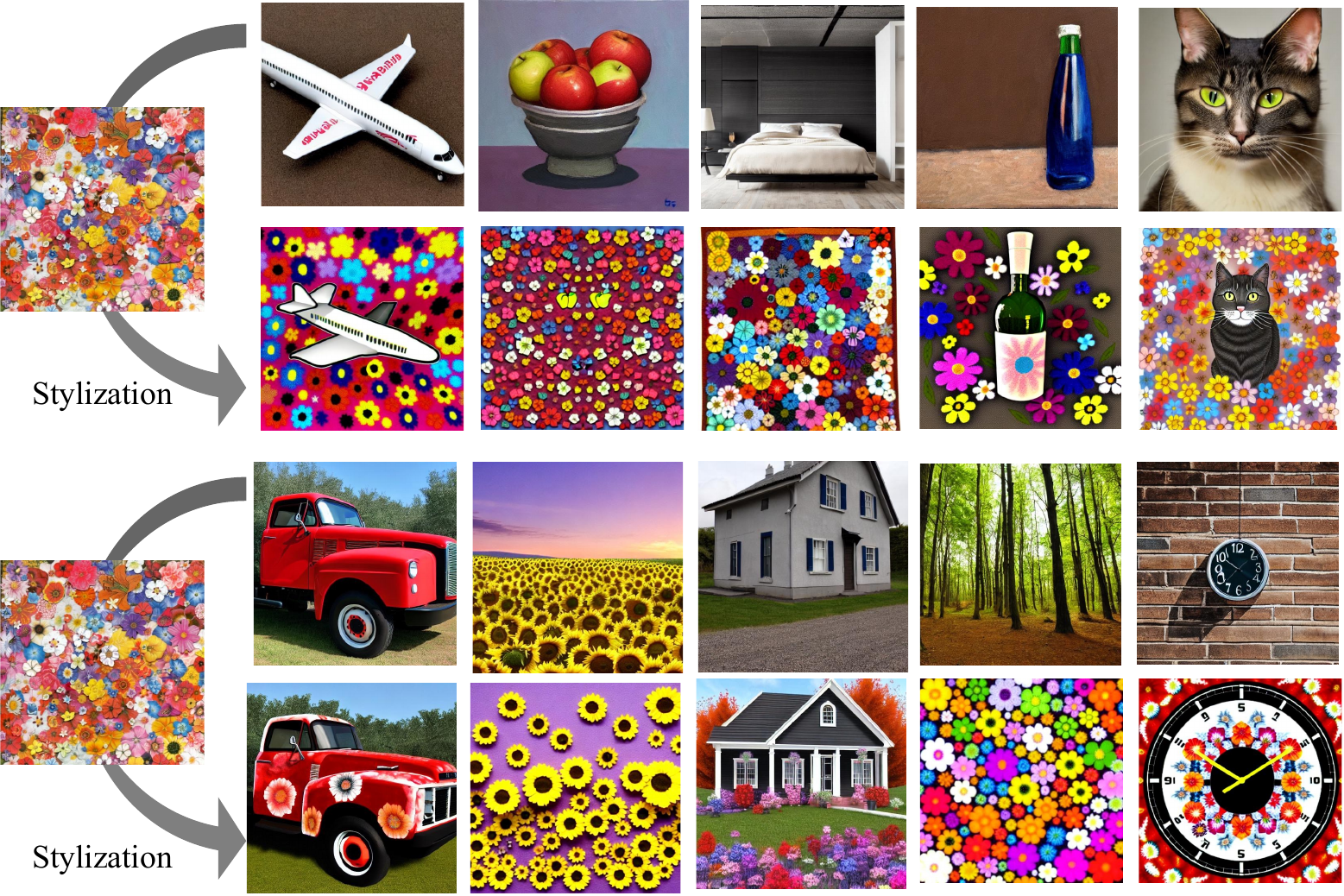}
    \caption{Stylization examples generated by DreamBooth.}
    \end{subfigure} 
    \caption{
    % In Textual Inversion, we generated the original images by using the prompt ``A photo of a <object name>'' and generated the style images by using the prompt ``A photo of a <object name> in the style of <bengiles>'', where <bengiles> is our learned text embedding. In DreamBooth, we generated the original images by using the prompt ``A <object name>'' and generated the style images by using the prompt ``A <object name> in the style of bengiles flowers'', where bengiles flowers is our input instance prompt. The original images and their corresponding stylized images use the same initial noises.
    Comparison to Textual-Inversion and DreamBooth for stylization.}
    \label{fig:text_inver}
\end{figure}
% \paragraph{Qualitative comparisions} 
We also provide visual comparisons to Textual Inversion~\cite{gal_image_2022} and DreamBooth~\cite{ruiz_dreambooth_2022} in Fig.~\ref{fig:text_inver}. 
We follow the implementation of Textual Inversion\footnote{\url{https://huggingface.co/docs/diffusers/training/text_inversion}} and DreamBooth\footnote{\url{https://huggingface.co/docs/diffusers/training/dreambooth}}. 
For textual inversion, we use the same target style image in Section \ref{subsec:style} as the training dataset. As in our AdjointDPM model, we use one style image for training, for fair comparison, the training dataset to Textual Inversion and DreamBooth also include only one style image. We set the \emph{learnable property} as ``style'', \emph{placeholder token} as ``<bengiles>'', \emph{initializer token} as ``flowers'' in Textual Inversion. Then we run 5000 epochs with a learning rate $5\times 10^{-4}$ to train the Textual Inversion. For DreamBooth, we set \emph{instance prompt} as ``bengiles flowers''. Then we run 1000 epochs with a learning rate $1\times 10^{-6}$ to train the DreamBooth. In Fig.~\ref{fig:text_inver}, we show the stylization examples generated by using Textual Inversion and DreamBooth. We can observe distinct differences in the stylization outcomes when comparing the Textual Inversion and DreamBooth approaches to our AdjointDPM methods.
In some cases, for Textual Inversion and DreamBooth, we have noticed that the main objects within an image can vanish, resulting in the entire image being predominantly occupied by the applied "style."

\section{Implementation of AdjointDPM}
\label{apdx:implementation}

In this section, we present the explicit AdjointDPM algorithm. For VP-SDE, we have $f(t) = \frac{\rmd \log\alpha}{\rmd t}$ and $g^2(t)=\frac{\rmd \sigma_t^2}{\rmd t}-2\frac{\rmd \log \alpha}{\rmd t}\sigma_t^2$. Based on the definition of $\by_t$ and $\rho$,  we obtain
\begin{align*}
    \by_t = \frac{\alpha_0}{\alpha_t} \bx_t, \quad \rho = \gamma(t) = \alpha_0\frac{\sigma_t}{\alpha_t}-\sigma_0. 
\end{align*}
We denote the timesteps for solving forward generation ODE as $\{t_i\}_{i=0}^N$, where $N$ is the number of timesteps.  Then based on this re-parameterization, we can show our explicit forward generation algorithm and reverse algorithm of obtaining gradients for VP-SDE in Algorithm~\ref{alg:node_forward} and Algorithm~\ref{alg:node_vpsde}. 

% \paragraph{Implementation Details of AdjointDPM}
For the choice of $\alpha_t$, $\sigma_t$, and sampling steps $\{t_i\}_{i=1}^N$, we adopt the implementation of DPM-solver\footnote{\url{https://github.com/LuChengTHU/dpm-solver}}. Specifically, we consider three options for the schedule of $\alpha_t$ and $\sigma_t$, discrete, linear, and cosine. The detailed formulas for obtaining $\alpha_t$ and $\sigma_t$ for each schedule choice are provided in~\cite[Appendix D.4]{lu2022dpm}. The choice of schedule depends on the specific applications.  We usually solve the forward generation ODE function from time $T$ to time $\epsilon$ ($\epsilon>0$ is a hyperparameter near 0). Regarding the selection of discrete timesteps $\{t_i\}_{i=1}^N$ in numerically solving ODEs, we generally divide the time range $[T, \epsilon]$ using one of three approaches: uniform, logSNR, or quadratic. The specific time splitting methods can be found in the DPM-solver. Subsequently, we obtain the generated images and gradients by following Algorithm~\ref{alg:node_forward} and Algorithm~\ref{alg:node_vpsde}. To solve ODE functions in these algorithms, we directly employ the \emph{odeint adjoint} function in the \emph{torchdiffeq} packages\footnote{\url{https://github.com/rtqichen/torchdiffeq}}. 

\begin{algorithm}
\caption{Forward generation by solving an ODE initial value problem}
\label{alg:node_forward}
\textbf{Input:} model $\bepsilon_{\theta}$, timesteps $\{t_i\}_{i=0}^N$, initial value $\bx_{t_0}$.
\begin{algorithmic}
\Statex $\by_{t_0} \gets \bx_{t_0}$; \Comment{Re-parameterize $\bx_{t_0}$}
\Statex $\{\rho_i\}_{i=1}^N \gets \{\gamma(t_i)\}_{i=1}^N$; \Comment{Re-parameterize timesteps}
\Statex $\by_{t_N}= \mathrm{ODESolve}\left(\by_{t_0}, \{\rho_i\}_{i=1}^n, \bepsilon_{\theta}\left(\frac{\alpha_t}{\alpha_{t_0}}\by_{t}, \gamma^{-1}(t),c\right)\right)$
\Comment{Solve forward generation ODE}
\end{algorithmic}
\textbf{Return:} $\bx_{t_N}=\frac{\alpha_{t_N}}{\alpha_{t_0}}\by_{t_N}$ 
\end{algorithm}

\begin{algorithm}
\caption{Reverse-mode derivative of an ODE initial value problem}
\label{alg:node_vpsde}
\textbf{Input:} model $\bepsilon_{\theta}$, timesteps $\{\rho_i\}_{i=1}^N$, final state $\by_{\rho_N}$, loss gradient $\partial L/\partial \by_{\rho_N}$.
\begin{algorithmic}
\Statex $a(\rho_N)=\frac{\partial L}{\partial \by_{\rho_N}}$, $a_{\theta}(\rho_N)=\textbf{0}$, $z_0=[\by_{\rho_N}, a(\rho_N), a_{\theta}(\rho_N)]$ \Comment{Define initial augmented state.}
\Statex \textbf{def} AugDynamics($[\by_{\rho}, \ba_{\rho}, \cdot], \rho,\theta$) \Comment{Define dynamics on augmented state.}\\
        $\qquad\textbf{return}$ $[\bs(\by_{\rho}, \rho, \theta, c), -\ba_{\rho}^T \frac{\partial \bs}{\partial \by}, -\ba_{\rho}^T \frac{\partial \bs}{\partial \theta}
        % -\ba_t^T \frac{\partial \bs}{\partial t}
        ]$ \Comment{Concatenate time-derivatives}

\Statex $[\by_{\rho_0},  \frac{\partial L}{\partial \by_{\rho_0}}, \frac{\partial L}{\partial \theta}
% \frac{\partial L}{\partial t_0}
]=\mathrm{ODESolve}(z_0,\mathrm{AugDynamics}, \{\rho_i\}_{i=1}^N, \theta )$ \Comment{Solve reverse-time ODE}
\end{algorithmic}
\textbf{Return:} $[\frac{\partial L}{\partial \bx_{t_0}}, \frac{\partial L}{\partial \theta}]$ \Comment{Return gradients}
\end{algorithm}

\end{document}

%% file: preamble.tex
 \usepackage[mathscr]{eucal}
 \usepackage[cmex10]{amsmath}
 \usepackage{amssymb,amsmath,amsthm,amsfonts,latexsym}
 \usepackage{fixltx2e}%ordering of single and double column floats
 \usepackage{array}%array and tabular environments
 \usepackage{verbatim}
 \usepackage{bm}
 \usepackage{bbm}
\usepackage{algorithm}
\usepackage{algpseudocode}
 \usepackage{verbatim}
 \usepackage{textcomp}
 \usepackage{mathrsfs}
 \usepackage{epstopdf}
 \usepackage{setspace}

\newcommand{\calN}{\mathcal{N}}

\newcommand{\ba}{\mathbf{a}}

\newcommand{\bF}{\mathbf{F}}

\newcommand{\bG}{\mathbf{G}}

\newcommand{\bI}{\mathbf{I}}

\newcommand{\bs}{\mathbf{s}}

\newcommand{\bw}{\mathbf{w}}

\newcommand{\bx}{\mathbf{x}}

\newcommand{\by}{\mathbf{y}}

\newcommand{\rmd}{\mathrm{d}}

\DeclareMathAlphabet{\mathbsf}{OT1}{cmss}{bx}{n}
\DeclareMathAlphabet{\mathssf}{OT1}{cmss}{m}{sl}% slanted sans serif

\DeclareSymbolFont{bsfletters}{OT1}{cmss}{bx}{n}  
\DeclareSymbolFont{ssfletters}{OT1}{cmss}{m}{n}
\DeclareMathSymbol{\bsfGamma}{0}{bsfletters}{'000}
\DeclareMathSymbol{\ssfGamma}{0}{ssfletters}{'000}
\DeclareMathSymbol{\bsfDelta}{0}{bsfletters}{'001}
\DeclareMathSymbol{\ssfDelta}{0}{ssfletters}{'001}
\DeclareMathSymbol{\bsfTheta}{0}{bsfletters}{'002}
\DeclareMathSymbol{\ssfTheta}{0}{ssfletters}{'002}
\DeclareMathSymbol{\bsfLambda}{0}{bsfletters}{'003}
\DeclareMathSymbol{\ssfLambda}{0}{ssfletters}{'003}
\DeclareMathSymbol{\bsfXi}{0}{bsfletters}{'004}
\DeclareMathSymbol{\ssfXi}{0}{ssfletters}{'004}
\DeclareMathSymbol{\bsfPi}{0}{bsfletters}{'005}
\DeclareMathSymbol{\ssfPi}{0}{ssfletters}{'005}
\DeclareMathSymbol{\bsfSigma}{0}{bsfletters}{'006}
\DeclareMathSymbol{\ssfSigma}{0}{ssfletters}{'006}
\DeclareMathSymbol{\bsfUpsilon}{0}{bsfletters}{'007}
\DeclareMathSymbol{\ssfUpsilon}{0}{ssfletters}{'007}
\DeclareMathSymbol{\bsfPhi}{0}{bsfletters}{'010}
\DeclareMathSymbol{\ssfPhi}{0}{ssfletters}{'010}
\DeclareMathSymbol{\bsfPsi}{0}{bsfletters}{'011}
\DeclareMathSymbol{\ssfPsi}{0}{ssfletters}{'011}
\DeclareMathSymbol{\bsfOmega}{0}{bsfletters}{'012}
\DeclareMathSymbol{\ssfOmega}{0}{ssfletters}{'012}

\newcommand{\bepsilon}{\bm{\epsilon}}

\usepackage{thmtools, thm-restate}